\documentclass[10pt,twocolumn,letterpaper]{article}

\usepackage{cvpr}
\usepackage{times}
\usepackage{epsfig}
\usepackage{graphicx}
\usepackage{amsmath}
\usepackage{amssymb}

\usepackage{mathtools}
\usepackage{multirow}
\usepackage{graphics}
\usepackage{subfig}
\usepackage{float}
\usepackage{stfloats}
\graphicspath{ {figures/} }

\usepackage[ruled,vlined]{algorithm2e}
\DeclarePairedDelimiter{\norm}{\lVert}{\rVert}

\usepackage[pagebackref=true,breaklinks=true,letterpaper=true,colorlinks,bookmarks=false]{hyperref}

\cvprfinalcopy 

\def\cvprPaperID{5888} 

\ifcvprfinal\fi
\begin{document}
\newcommand{\name}{VOLDOR}

\newcommand{\xiTheta}[2]{\boldsymbol{\xi_{#1}}\boldsymbol{(}\theta^{#2}\boldsymbol{)}}
\newcommand{\piTheta}[2]{\boldsymbol{\pi_{#1}}\boldsymbol{(}\theta^{#2}\boldsymbol{)}}
\newcommand{\flowTJ}{{\boldsymbol{X}_{\boldsymbol{t}}^{\piTheta{t\!-\!1}{j}}}}
\newcommand{\piX}[2]{\boldsymbol{\pi_{#1}}\boldsymbol{(} {#2} \boldsymbol{)}}
\newcommand{\flowTX}[1]{{\boldsymbol{X}_{\boldsymbol{t}}^{ \piX{t\!-\!1}{#1} }} }

\title{
VOLDOR: Visual Odometry from Log-logistic Dense Optical flow Residuals 
}
\author{Zhixiang Min\qquad\:Yiding Yang\qquad\:Enrique Dunn\\
Stevens Institute of Technology\\
{\tt\small \{zmin1,yyang99,edunn\}@stevens.edu}
}

\maketitle

\begin{abstract}
We propose a dense indirect visual odometry method taking as input externally estimated optical flow fields instead of hand-crafted feature correspondences.
We define our problem as a probabilistic model and develop a generalized-EM formulation for the joint inference of camera motion, pixel depth, and motion-track confidence.
Contrary to traditional methods assuming Gaussian-distributed observation errors, we supervise our inference framework under an (empirically validated) adaptive log-logistic distribution model.
Moreover, the log-logistic residual model generalizes well to different state-of-the-art optical flow methods, making our approach modular and agnostic to the choice of optical flow estimators. 
Our method achieved top-ranking results on both 
TUM RGB-D and KITTI odometry benchmarks. 
Our open-sourced implementation 
\footnote{\href{https://github.com/htkseason/VOLDOR}{{https://github.com/htkseason/VOLDOR}}} 
is inherently GPU-friendly with only linear computational and storage growth. 

\vspace{-0.04cm}

\end{abstract}

\section{Introduction}

Visual odometry (VO) \cite{nister2006visual, fraundorfer2011visual, fraundorfer2012visual} 
addresses 
the recovery of camera poses from an input video sequence, which supports applications such as augmented reality, robotics and autonomous driving. Traditional indirect VO \cite{kitt2010visual, pire2015stereo, mur2017orb} methods rely on the geometric analysis of sparse keypoint correspondences to determine multi-view relationships among the input video frames. Moreover, by virtue of relying on local feature detection and correspondence pre-processing modules, indirect methods pose the VO problem as a reprojection-error minimization task. Conversely, direct methods \cite{6126513, engel2014lsd, kerl2013dense}, strive to jointly determine a (semi-)dense registration (warping) across images, as well as the parameters of a camera motion model. By virtue of evaluating a dense correspondence field, direct methods strive to minimize photometric error among registered images. While both of these contrasting approaches have been successful in practice, there are important limitations to be addressed. An open problem within indirect methods, is how to characterize feature localization error within the context of VO \cite{kanazawa2003we, kanatani2004uncertainty, kanatani2008statistical, zeisl2009estimation}, where motion blur, depth occlusions and viewpoint variations may corrupt such estimates. Nevertheless, least squares methods are commonly used under the assumption of zero-mean Gaussian distributed observation errors. On the other hand, the efficacy of direct methods relies on the strict adherence to the small-motion and appearance-constancy assumptions (or on the development of registration models robust to such variations), which speaks to the difficulty of adequately modeling data variability in this context and, in turn, reduces the scope of their applicability. 

\begin{figure}[t]
\centering
\includegraphics[width=\columnwidth]{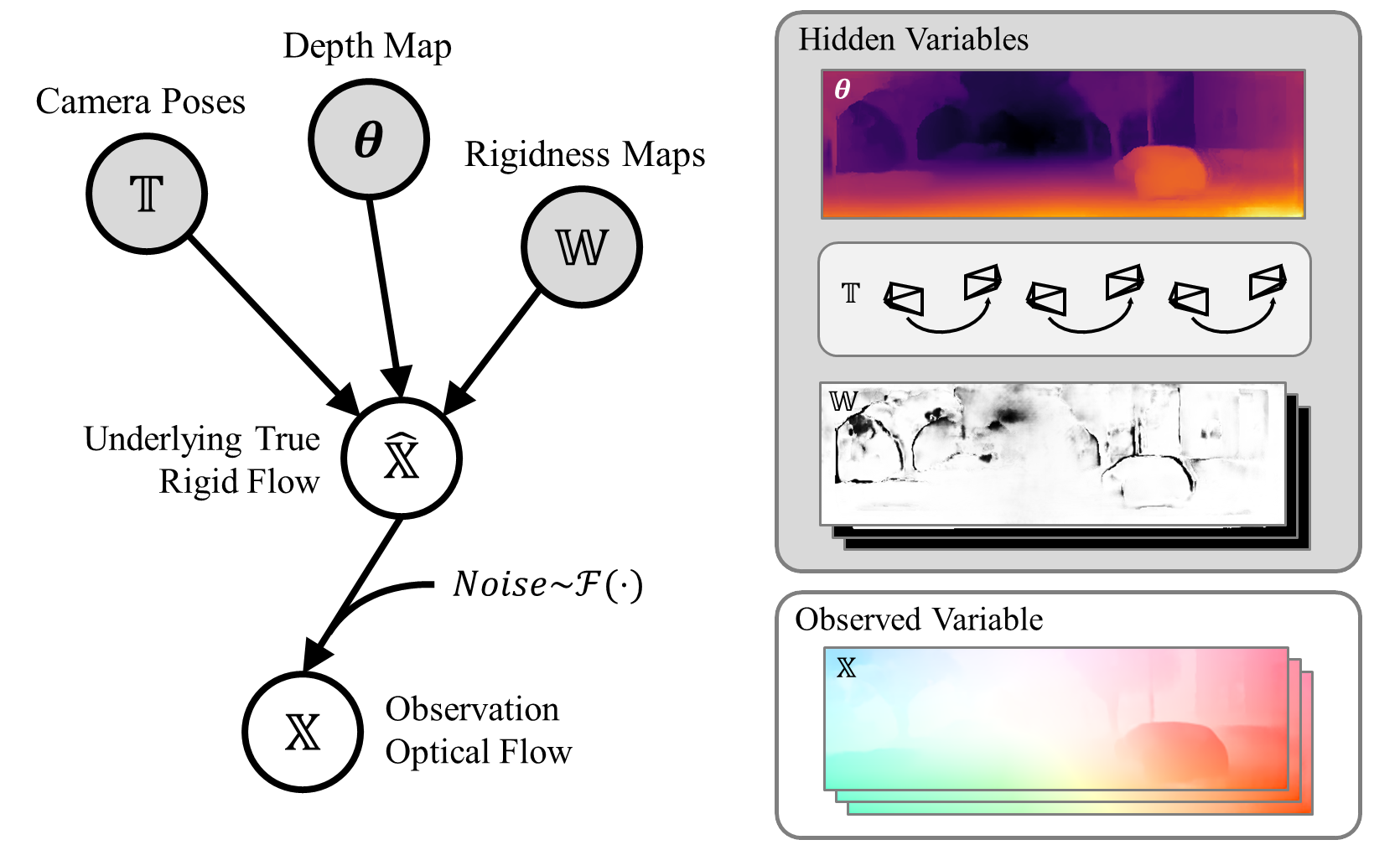}
\caption{\small  \textbf {\name{} probabilistic graphical model}. Optical flow field sequences are modeled as observed variables subject to Fisk-distributed measurement errors. Camera poses, depth map and rigidness maps are modeled as hidden variables.}
\label{fig:bayes_network}
\end{figure}

Recent developments on optical flow estimation \cite{sun2018pwc, ilg2017flownet} using supervised learning have yielded state-of-the-art performance. However, such performance benefits have not yet permeated to pose-estimation tasks, where standard multi-view geometry methods still provide the ``gold standard''. This work develops a dense indirect framework for monocular VO that takes as input externally computed optical flow from supervised-learning estimators. We have empirically observed that optical flow residuals tend to conform to a log-logistic (i.e. Fisk) distribution model parametrized by the optical flow magnitude. We leverage this insight to propose a probabilistic framework that fuses a dense optical flow sequence and jointly estimates camera motion, pixel depth, and motion-track confidence through a generalized-EM formulation. Our approach is dense in the sense that each pixel corresponds to an instance of our estimated random variables; it is indirect in the sense that we treat individual pixels as viewing rays within minimal feature-based multi-view geometry models (i.e. P3P for camera pose, 3D triangulation for pixel depth) and implicitly optimize for reprojection error. Starting from a deterministic bootstrap of a camera pose and pixel depths attained from optical flow inputs, we iteratively alternate the inference of depth, pose and track confidence over a batch of consecutive images.

The advantages of our framework include: 1) We present a modular framework that is agnostic to the optical flow estimator engine, which allows us to take full advantage of recent deep-learning optical flow methods. Moreover, by replacing sparse hand-crafted features inputs by learned dense optical flows, we gain surface information for poorly textured (i.e. feature-less) regions. 2) By leveraging our empirically validated log-logistic residual model, we attain highly accurate probabilistic estimates for our scene depth and camera motion, which do not rely on Gaussian error assumptions. Experiments on the KITTI \cite{geiger2012we} and TUM RGB-D \cite{sturm2012benchmark} benchmark have yielded top-ranking performance both on the visual odometry and depth estimation tasks. Our highly parallelizable approach also allows real-time application on commodity GPU-based architectures. 

\section{Related Works}

\textbf{Indirect methods.} Indirect methods \cite{yokozuka2019vitamin, huang2019clusterslam, nister2004efficient, forster2014svo, davison2007monoslam, gomez2016robust} rely on geometric analysis of sparse keypoint correspondences among input video frames, and pose VO problem as a reprojection-error minimization task. VISO \cite{kitt2010visual} employs Kalman filter with RANSAC-based outlier rejection to robustly estimate frame-to-frame motion. PTAM \cite{pire2015stereo} splits tracking and mapping to different threads, and applies expensive bundle adjustment (BA) at the back-end to achieve better accuracy while retaining real-time application. ORB-SLAM \cite{mur2015orb, mur2017orb} further introduces a versatile SLAM system with a more powerful back-end with global relocalization and loop closing, allowing large environment applications.

\textbf{Direct methods.} Direct methods \cite{pascoe2017nid, schubert2018direct, zhao2018good, kim2018linear, kerl2013dense, kerl2015dense, schops2019bad, whelan2015elasticfusion} maintain a (semi-)dense model, and estimates the camera motion by finding a warping that minimizes the photometric error w.r.t. video frames. DTAM \cite{6126513} introduces a GPU-based real-time dense modelling and tracking approach for small workspaces. LSD-SLAM \cite{engel2014lsd} switched to semi-dense model that allows large scale CPU real-time application. DSO \cite{engel2017direct} builds sparse models and combines a probabilistic model that jointly optimizes for all parameters as well as further integrates a full photometric calibration to achieve current state-of-the-art accuracy.

\textbf{Deep learning VO.} Recently, deep-learning has shown thriving progress on the visual odometry problem.
Boosting VO through geometric priors from learning-based depth predictions has been presented in \cite{yang2018deep, loo2019cnn, tateno2017cnn}.
Integration of deep-representations into components such as feature-points, depth maps and optimizers has been presented in \cite{tang2018ba, bloesch2018codeslam, clark2018ls, costante2015exploring}.
Deep-learning framework that jointly estimate depth, optical flow and camera motion has been presented in \cite{zhou2017unsupervised, yin2018geonet, zhan2018unsupervised, ummenhofer2017demon, wang2018learning, mahjourian2018unsupervised, zou2018df, ilg2018occlusions, wang2019unos, ranjan2019competitive, sheng2019unsupervised, bozorgtabar2019syndemo, chen2019self, zhao2018learning}. 
Further adding recurrent neural networks for learning temporal information is presented in \cite{wang2019recurrent, wang2017deepvo, xue2019beyond, li2019sequential}. However, deep learning methods are usually less explainable and have difficulty when transferring to unseen dataset or cameras with different calibration. Moreover, the precision of such methods still underperforms the state of the art.

\textbf{Deep learning optical flow.} Contrary to learning-based monocular depth approaches \cite{fu2018deep, godard2017unsupervised, godard2019digging, zhang2019pattern, guo2018learning}, which develop and impose strong semantic priors, learning for optical flow estimation may be informed by photometric error and achieve better generalization. Recent deep learning works on optical flow estimation \cite{wang2018occlusion,hui2018liteflownet, yang2018conditional, ranjan2017optical, janai2018unsupervised, liu2019selflow, hur2019iterative, zhong2019unsupervised, ilg2017flownet, dosovitskiy2015flownet, sun2018pwc} have shown satisfying accuracy, robustness and generalization,  outperforming traditional methods, especially under challenging conditions such as texture-less regions, motion blur and large occlusions. FlowNet \cite{dosovitskiy2015flownet}  introduced an encoder-decoder convolutional neural network for optical flow.  FlowNet2 \cite{ilg2017flownet} improved its performance by stacking multiple basic FlowNets. Most recently, PWC-Net \cite{sun2018pwc} integrates  spatial pyramids, warping and cost volumes into deep optical flow estimation, improving  performance and generalization to the current state-of-the-art.

\begin{figure*}[t]
\centering
\includegraphics[width=2\columnwidth]{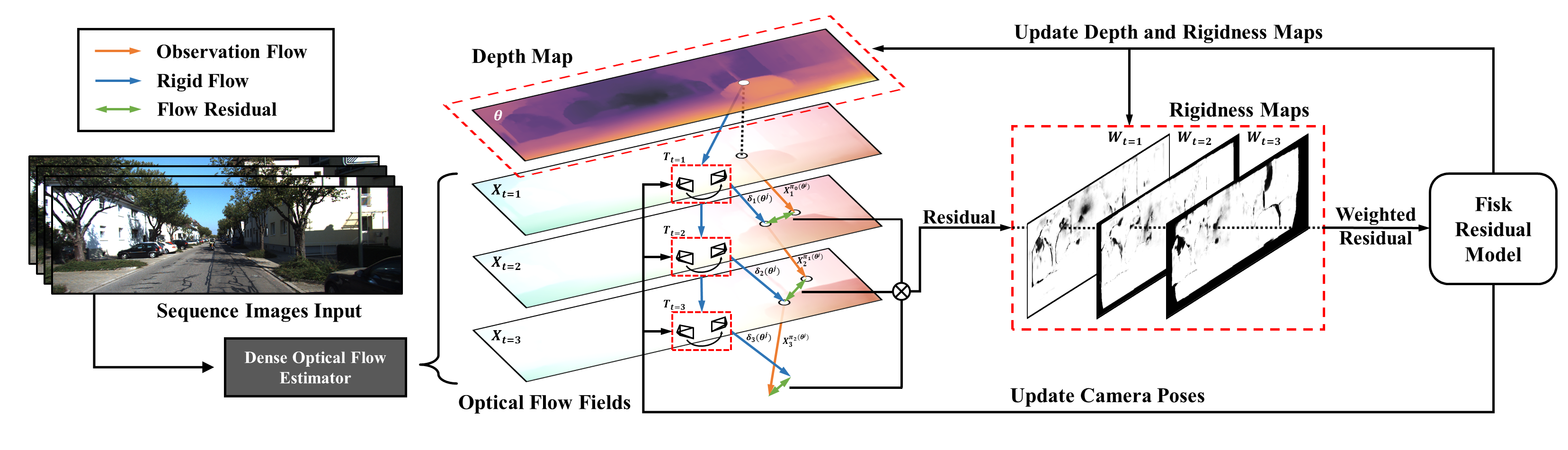}
\caption{\small  \textbf{Iterative estimation workflow}. We input externally computed optical flow estimates of a video sequence. Scene depth, camera poses and a rigidness maps are alternatively estimated by enforcing congruence between predicted rigid flow and input flow observations. Estimation is posed as a probabilistic inference task governed by a Fisk-distributed residual model.}
\label{fig:workflow}
\end{figure*}

\section{Problem Formulation} \label{sec:problem_formulation}
{\bf Rationale}. 
Optical flow can been seen as a combination of rigid flow, which is relevant to the camera motion and scene structure, along with an unconstrained flow describing general object motion \cite{wulff2017optical}. Our VO method inputs a batch of externally computed optical flow fields and infers the underlying temporally-consistent aforementioned scene structure (depth map), camera motions, as well as pixel-wise probabilities for the ``rigidness" of each optical flow estimate. Further, we posit the system framework under the supervision of an empirically validated adaptive log-logistic residual model over the end-point-error (EPE) between the estimated rigid flow and the input (observed) flow.


{\bf Geometric Notation}. 
We input a sequence of externally computed (observed) dense optical flow fields $\mathbb{X}=\{ \boldsymbol{X_t} \mid t=1, \cdots, t_N \}$, where $\boldsymbol{X_t}$ is the optical flow map from image $\boldsymbol{I_{t-1}}$ to $\boldsymbol{I_t}$, while $\boldsymbol{X_t^j}=(u_t^j,v_t^j)^T$ denotes the  optical flow vector of pixel $j$ at time $t$. We aim to infer the camera poses  $\mathbb{T}=\{ \boldsymbol{T_t} \mid t=1, \cdots, t_N \}$, where $\boldsymbol{T_t} \in SE(3)$ represents the relative motion from time $t-1$ to $t$.\\
\noindent 
To define a likelihood model  relating our observations $\mathbb{X}$ to $\mathbb{T}$, we introduce two additional (latent) variable types: 1) a depth field $\boldsymbol{\theta}$ defined over $\boldsymbol{I_0}$; where we denote $\theta^j$ as the depth value at pixel $j$, and 2) a rigidness probability map $\boldsymbol{W_t}$ associated to $\boldsymbol{\theta}$ at time $t$;  while $\mathbb{W}=\{\boldsymbol{W_t} \mid t=1, \cdots, t_N \}$ denotes the set of rigidness maps, and $W_t^j$ denotes the rigidness probability of pixel $j$ at time $t$. 
\\
\noindent Having depth map $\boldsymbol{\theta}$ and rigidness maps $\mathbb{W}$, we can obtain a rigid flow $\xiTheta{t}{j}$ by applying the rigid transformation $\mathbb{T}$ to the point cloud associated with $\boldsymbol{\theta}$, conditioned on $\mathbb{W}$. Assuming $\boldsymbol{T_0}=\boldsymbol{I}$, we let $\piTheta{t}{j}$ denote the pixel coordinate of projecting the 3D point associated with $\theta^j$ into the camera image plane at time $t$ using given camera poses $\mathbb{T}$, by
\begin{equation}
\piTheta{t}{j}=
\boldsymbol{K}\left(\prod_{i=0}^t\boldsymbol{T_i}\right)\theta^j \boldsymbol{K}^{-1}[x_j\;y_j\;1]^T 
\end{equation}
where $\boldsymbol{K}$ is the camera intrinsic and $x_j, y_j$ are the image coordinates of pixel $j$. Hence, rigid flow can be defined as $\xiTheta{t}{j} = \piTheta{t}{j} - \piTheta{t-1}{j}$.

{\bf Mixture Likelihood Model}
We model the residuals between observation flow and rigid flow with respect to the continuous rigidness probability $W_t^j$.
\begin{equation} \label{eq:likelihood}
\begin{split}
P(\flowTJ & \mid \theta^j, \boldsymbol{T_t}, W_t^j  \,;\, \boldsymbol{T_{1}} \cdots \boldsymbol{T_{t-1}})
\\
&=\begin{cases}
\rho( \xiTheta{t}{j} \mid\mid \flowTJ ) & \text{if $W_t^j=1$}
\\
\mu( \flowTJ ) & \text{if $W_t^j=0$}
\end{cases}\\
\end{split}
\end{equation}
where the probability density function $\rho(\cdot\mid\mid\cdot)$ represents the probability for having the rigid flow $\xiTheta{t}{j}$ under the observation flow of $\,\flowTJ\,$, and $\mu(\cdot)$ is a uniform distribution whose density varies with $\,\flowTJ\,$. We will define these two functions in \S\ref{sec:likelihood_model}. Henceforth, when modeling the probability of $\boldsymbol{X_t}$, we only write down $\boldsymbol{T_t}$ in the conditional probability, although the projection also depends on preceding camera poses $\boldsymbol{T_{1}}, \cdots, \boldsymbol{T_{t-1}}$, that we assume fixed and for which $\boldsymbol{X_t}$ inherently does not contain any information. Moreover, jointly modeling for all previous camera poses along  with $\boldsymbol{X_t}$ would bias them  as well as increase the computational complexity. In the following paragraph, we will denote Eq. \eqref{eq:likelihood} simply as $P(\boldsymbol{X_t^{\pi_{t-1}(\theta^j)}} \mid \theta^j, \boldsymbol{T_t}, W_t^j)$. At this point, our visual odometry problem can be modeled as a maximum likelihood estimation problem
\begin{equation} \label{eq:MLE}
\begin{split}
&\underset{\boldsymbol{\theta}, \mathbb{T}, \mathbb{W}}{argmax}P(\mathbb{X} \mid \boldsymbol{\theta}, \mathbb{T}, \mathbb{W}) \\
=\;&\underset{\boldsymbol{\theta}, \mathbb{T}, \mathbb{W}}{argmax}
\prod_{t}\prod_{j}P(\flowTJ \mid \theta^j, \boldsymbol{T_t}, W_t^j)
\end{split}
\end{equation}
Furthermore, we promote spatial consistency among both our dense hidden variables $\theta$ and $\mathbb{W}$, through different mechanisms described in \S\ref{sec:depth_update}.

\section{Fisk Residual Model} \label{sec:likelihood_model}
\begin{figure}
\centering
\includegraphics[width=0.80\columnwidth]{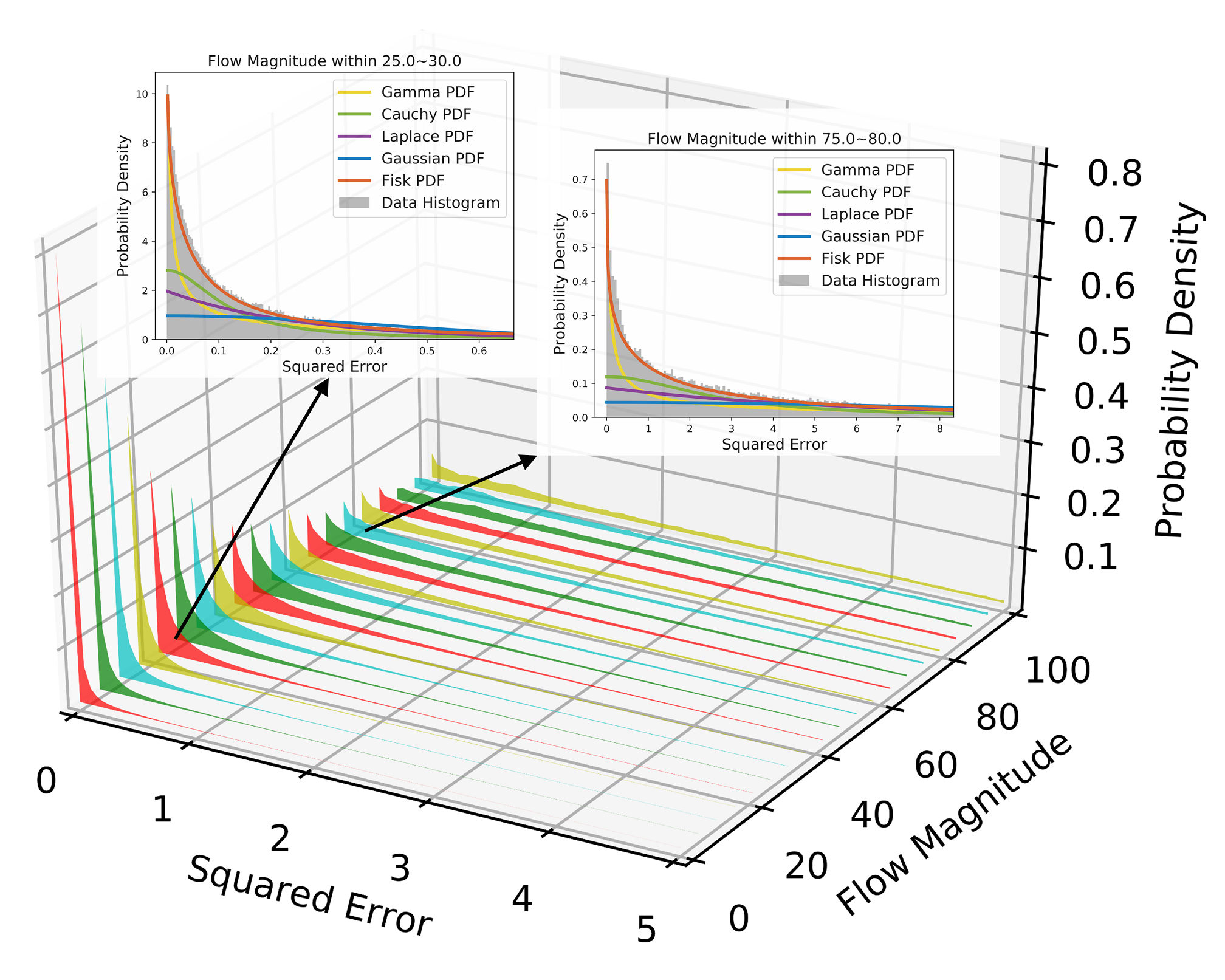} 
\caption{\small \textbf{Empirical residual distribution.} Optical flow EPE residual over flow magnitude for PWC-Net outputs on the entire groundtruth data for the KITTI \cite{geiger2012we} and Sintel \cite{Butler:ECCV:2012} datasets.}
\label{fig:fisk_1}
\end{figure}

The choice of an observation residual model plays a critical role for accurate statistical inference from a reduced number of observations \cite{gomez2017accurate, jaegle2016fast}, where in our case, a residual is defined as the end-point error in pixels between two optical flow vectors. 

\begin{figure}[]
\centering
\begin{tabular}{cc}
\subfloat[K-S Test]{\includegraphics[width=0.52\columnwidth]{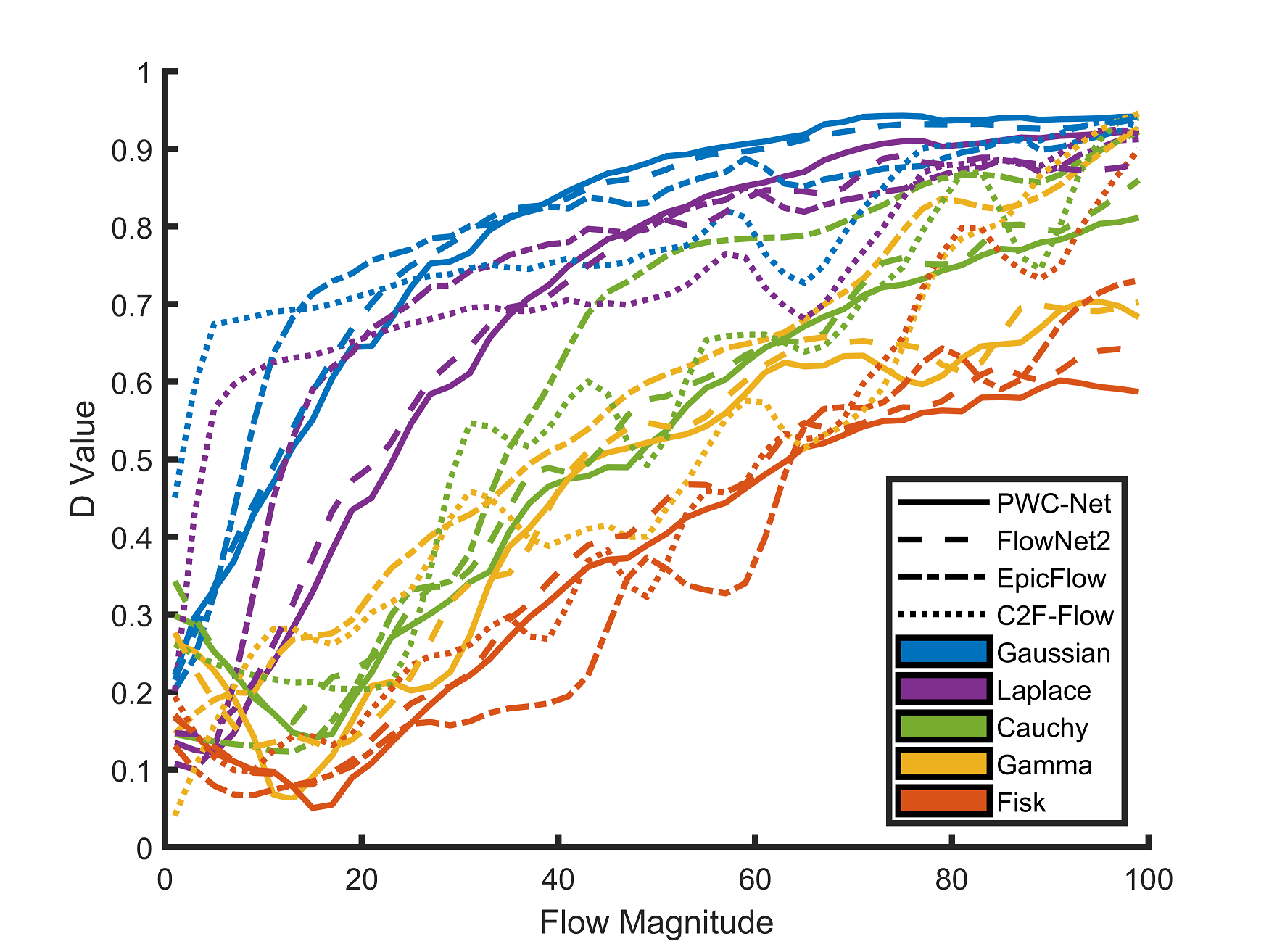}} & \subfloat[Parameterization of $\alpha, \beta$]{\includegraphics[width=0.38\columnwidth]{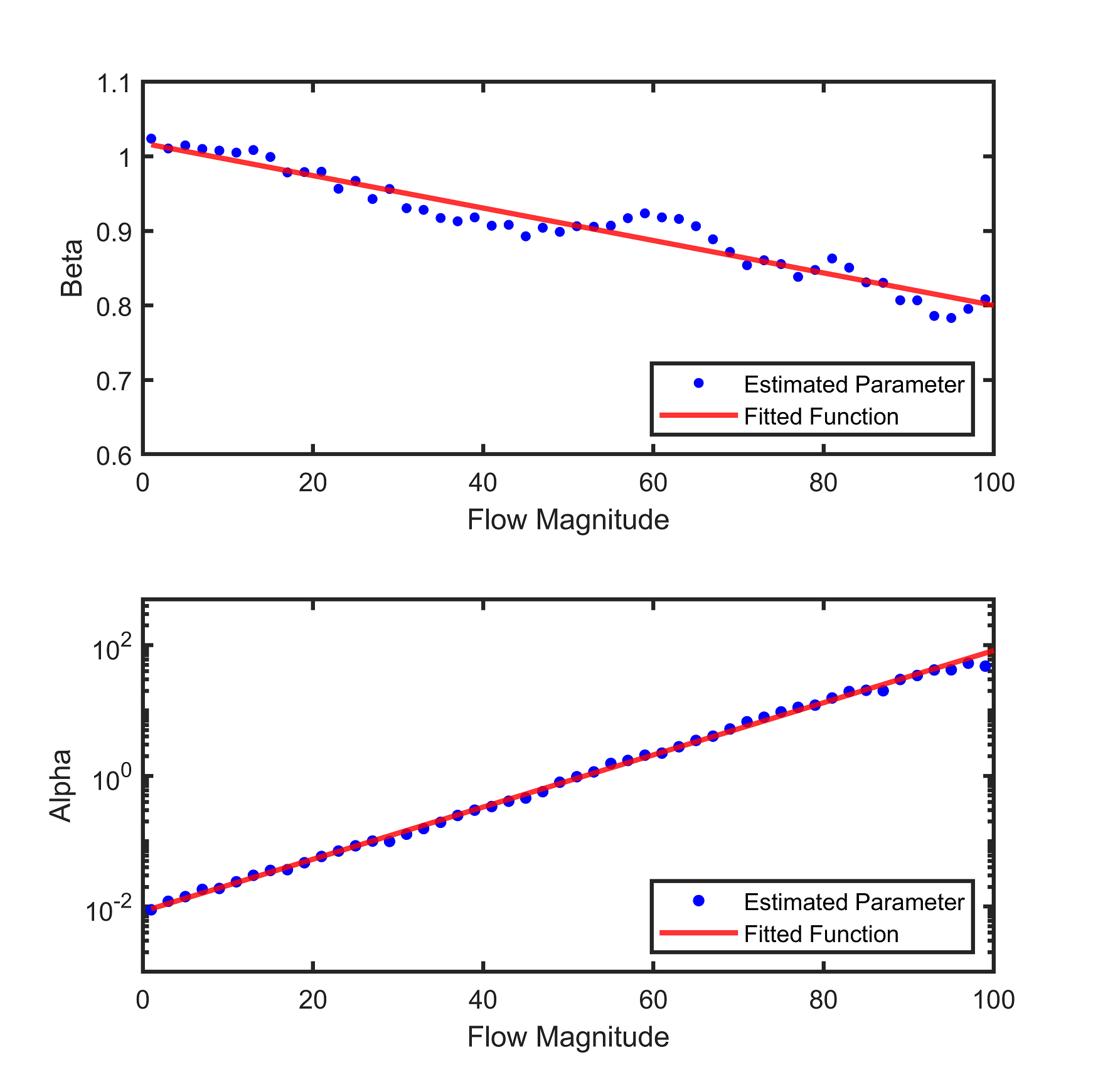}}
\end{tabular}
\caption{\small \textbf{Fitness quantification and model parameterization.} (a) Results of KS-test \cite{babu2004goodness} over four optical flow methods with five distributions (lower D values indicate better fit). (b) $\alpha,\beta$ are estimated from KITTI empirical data using, respectively, log-linear and linear regression, as described in Eq. \eqref{eq:fisk_param_fun_alpha}, \eqref{eq:fisk_param_fun_beta}.}
\label{fig:fisk_2}
\end{figure}

In practice, hierarchical optical flow methods (i.e. relying on recursive scale-space analysis) \cite{sun2018pwc, liu2009beyond} tend to amplify estimation errors in proportion to the magnitude of the pixel flow vector. In light of this, we explore an adaptive residual model determining the residuals distribution w.r.t. the magnitude of optical flow observations.
In Figure \ref{fig:fisk_1}, we empirically analyzed the residual distribution of multiple leading optical flow methods \cite{sun2018pwc, ilg2017flownet, liu2009beyond, revaud2015epicflow} w.r.t. groundtruth. 
We fit the empirical distribution to five different analytic models, and found the Fisk distribution to yield the most congruent shape over all flow magnitudes (See Figure \ref{fig:fisk_1} overlay). 
To quantify the goodness of fit, in Figure \ref{fig:fisk_2} (a), we posit K-S test results \cite{babu2004goodness}, which quantify the supremum distance (D value) of the CDF between our empirical distribution and a reference analytic distribution.

Hence, given  $\boldsymbol{v_{ob}} \!=\! \flowTJ$, we model the probability of  $\boldsymbol{v_{rig}} \!=\! \xiTheta{t}{j}$ matching the underlying groundtruth, as 
\begin{equation}
\rho(\boldsymbol{v_{rig}} \Vert \boldsymbol{v_{ob}}) = 
\mathcal{F}(\norm{\boldsymbol{v_{rig}}-\boldsymbol{v_{ob}}}_2^2; A(\boldsymbol{v_{ob}}), B(\boldsymbol{v_{ob}})),
\end{equation}
where the functional form for the PDF of the Fisk distribution $\mathcal{F}$ is given by
\begin{equation}
\mathcal{F}(x;\alpha,\beta) = \frac
{(\beta/\alpha)(x/\alpha)^{\beta-1}}
{(1+(x/\alpha)^\beta)^2}
\end{equation}
From Figure \ref{fig:fisk_2} (b), we further determine the parameters of Fisk distribution. Since a clear linear correspondence is shown in the figure, instead of using a look-up table, we applied the fitted function to find the parameters, as
\begin{equation} \label{eq:fisk_param_fun_alpha}
A(\boldsymbol{v_{ob}}) = a_1\boldsymbol{e}^{a_2\norm{\boldsymbol{v_{ob}}}_2}
\end{equation}
\begin{equation} \label{eq:fisk_param_fun_beta}
B(\boldsymbol{v_{ob}}) = {b_1}\norm{\boldsymbol{v_{ob}}}_2+b_2
\end{equation}
where $a_1, a_2$ and $b_1, b_2$ are learned parameters depending on the optical flow estimation method.

Next, we model the outlier likelihood function $\mu(\cdot)$. The general approach \cite{6746218, zheng2014patchmatch} is to assign outliers with uniform distribution to improve the robustness. In our work, for utilizing the prior given by the observation flow, we further let the density of uniform distribution be a function $\mu(\cdot)$ over the observation flow vector.
\begin{equation} \label{eq:adaptive_uniform}
\mu(\boldsymbol{v_{ob}})=
\mathcal{F}(\lambda^2\norm{\boldsymbol{v_{ob}}}_2^2; A(\boldsymbol{v_{ob}}), B(\boldsymbol{v_{ob}})))
\end{equation}
where $\lambda$ is a hyper-parameter adjusting the density, which is also the strictness for choosing inliers. The numerical interpretation of $\lambda$ is the optical flow percentage EPE for indistinguishable outlier ($W_t^j=0.5$). Hence, flows with different magnitudes can be compared under a fair metrics to be selected as an inlier.
\section{Inference}
In this section, we introduce our iterative inference framework which alternately optimizes depth map, camera poses and rigidness maps.

\subsection{Depth and Rigidness Update} \label{sec:depth_update}
\textbf{Generalized Expectation-Maximization (GEM).}
We infer depth $\boldsymbol{\theta}$ and its rigidness $\mathbb{W}$ over time while assuming fixed known camera pose $\mathbb{T}$. We approximate the true posterior $P(\mathbb{X} \mid \boldsymbol{\theta}, \mathbb{W} \,;\, \mathbb{T})$ through a GEM framework \cite{zheng2014patchmatch}. In this section, we will denote Eq. \eqref{eq:likelihood} as $P(\mathbb{X} \mid \boldsymbol{\theta}, \mathbb{W})$,  where the fixed $\mathbb{T}$ is omitted. 
We approximate the intractable real posterior $P(\boldsymbol{\theta}, \mathbb{W} \mid \mathbb{X})$ with a restricted family of distributions $q(\boldsymbol{\theta}, \mathbb{W})$, where $q(\boldsymbol{\theta}, \mathbb{W}) = \prod_{j}q(\theta^j) \prod_{t} q(\boldsymbol{W_t})$. For tractability, $q(\theta^j)$ is further constrained to the family of Kronecker delta functions $q(\theta^j) = \delta(\theta^j=\theta^{j*})$, where $\theta^{j*}$ is a parameter to be estimated. Moreover, $q(\boldsymbol{W_t})$ inherits the smoothness defined on the rigidness map $\boldsymbol{W_t}$ in Eq. \eqref{eq:smoothness}, which is proved in \cite{zheng2014patchmatch} as minimizing KL divergence between the variational distribution and the true posterior.
In the M step, we seek to estimate an optimal value for $\theta^j$ given an estimated PDF on $W_t^j$. Next, we describe our choice for the estimators used for this task.

{\bf Maximimum Likelihood Estimator (MLE).}
The standard definition of the MLE for our problem is given by
\begin{equation}
\resizebox{0.9\columnwidth}{!}{
$
\theta^j_{\scalebox{0.6}{MLE}} \!=\! \underset{\theta^{j*}}{argmax} \sum_t q(W_t^j) \log P(\flowTJ \mid \theta^j \!=\! \theta^{j*}, W_t^j)
$
}
\end{equation}
where $q(W_t^j)$ is the estimated distribution density given by E step. However, we empirically found the MLE criteria to be overly sensitive to inaccurate initialization.  More specifically, we bootstrap our depth map using only the first optical flow and use its depth values to sequentially bootstrap subsequent camera poses (more details in \S\ref{sec:pose_update} and \S\ref{sec:algorithm_integration}). Hence, for noisy/inaccurate initialization, using MLE for estimate refinement will impose high selectivity pressure on the rigidness probabilities $\mathbb{W}$, favoring a reduced number of higher-accuracy initialization. Given the sequential nature of our image-batch analysis, this tends to effectively reduce the set of useful down-stream observations used to estimate subsequent cameras.


{\bf Maximum Inlier Estimation (MIE).}
To reduce the bias caused by initialization and sequential updating, we relax the MLE criteria to the following MIE criteria,
\begin{equation} \label{eq:depth_criteria}
\resizebox{\columnwidth}{!}{
$
\theta^j_{\scalebox{0.6}{MIE}} \!=\! \underset{\theta^{j*}}{argmax}  \sum_t q(W_t^j \!=\! 1)  \log \frac{P(\flowTJ \mid \theta^j=\theta^{j*}, W_t^j=1)} {\sum_{W_t^j} P(\flowTJ \mid \theta^j=\theta^{j*}, W_t^j)}
$
}
\end{equation}
which finds a depth maximizing for the rigidness (inlier selection) map $\mathbb{W}$. We provide experimental details regarding the MIE criteria in \S\ref{sec:ablation_exp}.

\begin{figure}[]
\centering
\includegraphics[width=\columnwidth]{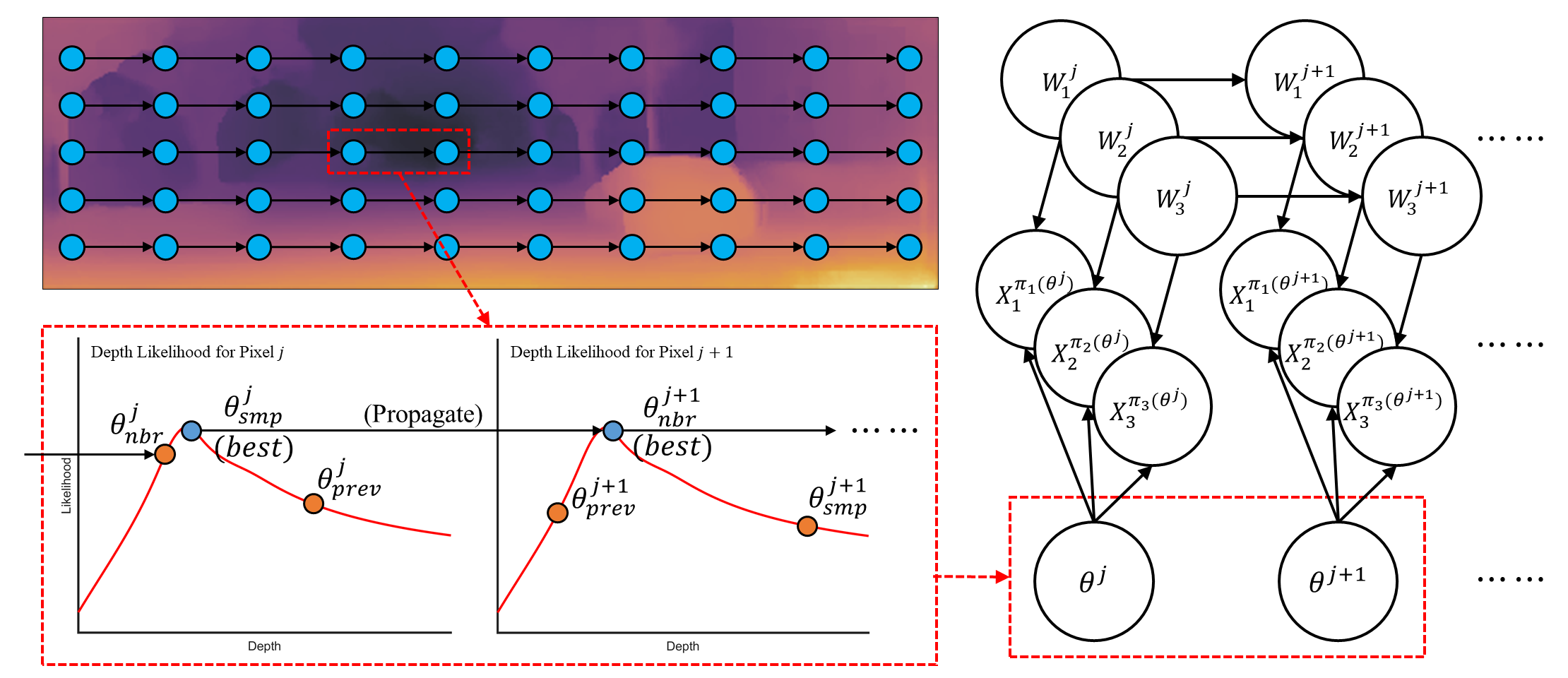}
\caption{\small  \textbf{Model for depth inference}. The image 2D field is broken into alternatively directed 1D chains, while depth values are propagated through each chain. Hidden Markov chain smoothing is imposed on the rigidness maps.}
\label{fig:workflow_infer_depth}
\end{figure}

We optimize $\theta_{\scalebox{0.6}{MIE}}^j$ through a sampling-propagation scheme as shown in Fig. \ref{fig:workflow_infer_depth}. A randomly sampled depth $\theta_{smp}^j$ is compared with the previous depth value $\theta_{prev}^j$ together with a value propagated from the previous neighbor $\theta^{j-1}_{nbr}$. Then, $\theta_{\scalebox{0.6}{MIE}}^j$ will be updated to the value of the best estimation among these three options. The updated $\theta^{j}_{\scalebox{0.6}{MIE}}$ will further be propagated to the neighbor pixel $j+1$.

{\bf Updating the Rigidness Maps.}
We adopt a scheme where the image is split into rows and columns, reducing a 2D image to several 1D hidden Markov chains, and a pairwise smoothness term is posed on the rigidness map
\begin{equation} \label{eq:smoothness}
P(W_t^j \mid W_t^{j-1})=
\begin{pmatrix}
\gamma & 1-\gamma
\\
1-\gamma & \gamma
\end{pmatrix}
\end{equation}
where $\gamma$ is a transition probability  encouraging similar neighboring rigidness.
In the E step, we update rigidness maps $\mathbb{W}$ according to $\boldsymbol{\theta}$. As the smoothness defined in Eq. \eqref{eq:smoothness}, the forward-backward algorithm is used for inferring $\mathbb{W}$ in the hidden Markov chain.
\begin{equation} \label{eq:rigidness_smooth}
q(W_t^j) = \frac{1}{A} m_f(W_t^j)m_b(W_t^j)
\end{equation}
where $A$ is a normalization factor while $m_f(W_t^j)$ and $m_b(W_t^j)$ are the forward and backward message of $W_t^j$ computed recursively as
\begin{equation}
m_f(W_t^j) = P_{ems}^{t,j} {\sum}_{W_t^{j\!-\!1}} m_f(W_t^{j-1}) P(W_t^j \mid W_t^{j-1})
\end{equation}
\begin{equation}
m_b(W_t^j) = {\sum}_{W_t^{j\!+\!1}} m_b(W_t^{j+1}) P_{ems}^{t,j+1} P(W_t^{j+1} \mid W_t^{j})
\end{equation}
where $P_{ems}^j$ is the emission probability refer to Eq. \eqref{eq:likelihood}, $P_{ems}^{t,j} = P(\flowTJ \mid \theta^j, W_t^j \,;\, \mathbb{T})$.

\subsection{Pose Update} \label{sec:pose_update}
We update camera poses while assuming fixed known depth $\boldsymbol{\theta}$ and rigidness maps $\mathbb{W}$. We use the optical flow chains in $\mathbb{X}$ to determine the 2D projection of any given 3D point extracted from our depth map. Since we aim to estimate relative camera motion we express scene depth relative to the camera pose at time $t-1$ and use the attained 3D-2D correspondences to define a dense PnP instance \cite{zheng2013revisiting}. We solve this instance by estimating the mode on an approximated posterior distribution given by Monte-Carlo sampling of the pose space through (minimal) P3P instances. 
The robustness to outliers correspondences and the independence from initialization provided by our method is crucial to
bootstrapping our visual odometry system \S\ref{sec:algorithm_integration}, where the camera pose need to be estimated from scratch with an uninformative rigidness map (all one). 

The problem can be written down as maximum a posterior (MAP) by
\begin{equation}
\underset{\mathbb{T}}{argmax}
P(\mathbb{T} \mid \mathbb{X} \,;\, \boldsymbol{\theta}, \mathbb{W})
\end{equation}
Finding the optimum camera pose is equal to compute the maximum posterior distribution $P(\mathbb{T} \mid \mathbb{X} \,;\, \boldsymbol{\theta}, \mathbb{W})$, which is not tractable since it requires to integral over $\mathbb{T}$ to compute $P(\mathbb{X})$. Thus, we use a Monte-Carlo based approximation, where for each depth map position $\theta^{j_1}$ we randomly sample two additional distinct positions $\{{j_2},{j_3}\}$ to form a 3-tuple $\boldsymbol{\Theta^g}=\{\theta^{j_1}, \theta^{j_2}, \theta^{j_3}\}$, with associated rigidness values $\mathbb{W}_t^g=\{W_t^{j_1},W_t^{j_2},W_t^{j_3}\}$, to represent the $g^{th}$ group. Then the posterior can be approximated as
\begin{equation}
\begin{split}
&P(\mathbb{T} \mid \mathbb{X} \,;\, \boldsymbol{\theta}, \mathbb{W})
\\
\approx & \prod_{t}
\left[
\frac{1}{S}\sum_{g}^{S}
P(\boldsymbol{T_t} \mid \flowTX{\boldsymbol{\Theta^g}} \,;\, \boldsymbol{\Theta^g}, \mathbb{W}_t^g) \right]
\end{split}
\end{equation}
where $S$ is the total number of groups. Although the posterior $P(\boldsymbol{T_t} \mid \flowTX{\boldsymbol{\Theta^g}} \,;\, \boldsymbol{\Theta^g}, \mathbb{W}_t^g)$ is still not tractable, using 3 pairs of 3D-2D correspondences, PnP reaches its minimal form of P3P, which can be solved efficiently using the
P3P algorithm \cite{gao2003complete, kneip2011novel, masselli2014new, ke2017efficient}. Hence we have
\begin{equation}
\resizebox{0.95\columnwidth}{!}{
$
\boldsymbol{\hat{T}_t^g} = \phi (
\left(\prod_{i=0}^{t-1}\boldsymbol{T_i}\right) \boldsymbol{\Theta^g} \boldsymbol{K^{-1}}[x_g\;y_g\;1]^T,
\piX{t\!-\!1}{\boldsymbol{\Theta^g}} \!+\! \flowTX{\boldsymbol {\Theta^g}}
)
$
}
\end{equation}
where $\phi(\cdot, \cdot)$ denotes the P3P solver, for which we use AP3P \cite{ke2017efficient}. The first input argument indicates the 3D coordinates of selected depth map pixels at time $t-1$ obtained by combining previous camera poses, while the second input argument is their 2D correspondences at time $t$, obtained using optical flow displacement. Hence, we use a tractable variational distribution $q(\boldsymbol{T_t^g})$ to approximate its true posterior.
\begin{equation} \label{eq:pose_variational_distribution}
q(\boldsymbol{T_t^g}) \sim \mathcal{N}(\boldsymbol{\hat{T}_t^g}, \boldsymbol{\Sigma})
\end{equation}
where $q(\boldsymbol{T_t^g})$ is a normal distribution with mean $\boldsymbol{\hat{T}_t^g}$ and predefined fixed covariance matrix $\boldsymbol{\Sigma}$ for simplicity. Furthermore, we weight each variational distribution with $\norm{\mathbb{W}_t^g}=\prod_{W_i \in \mathbb{W}_t^g}W_i$, such that potential outliers indicated by rigidness maps can be excluded or down weighted. Then, the full posterior can be approximated as
\begin{equation} \label{eq:pose_criteria}
P(\mathbb{T} \mid \mathbb{X} \,;\, \boldsymbol{\theta}, \mathbb{W}) \approx
\prod_{t}
\left[
\frac{1}{\sum_g\norm{\mathbb{W}_t^g}}
\sum_{g}
\norm{\mathbb{W}_t^g}q(\boldsymbol{T_t^g})
\right]
\end{equation}

\begin{figure}[]
\centering
\includegraphics[width=0.9\columnwidth]{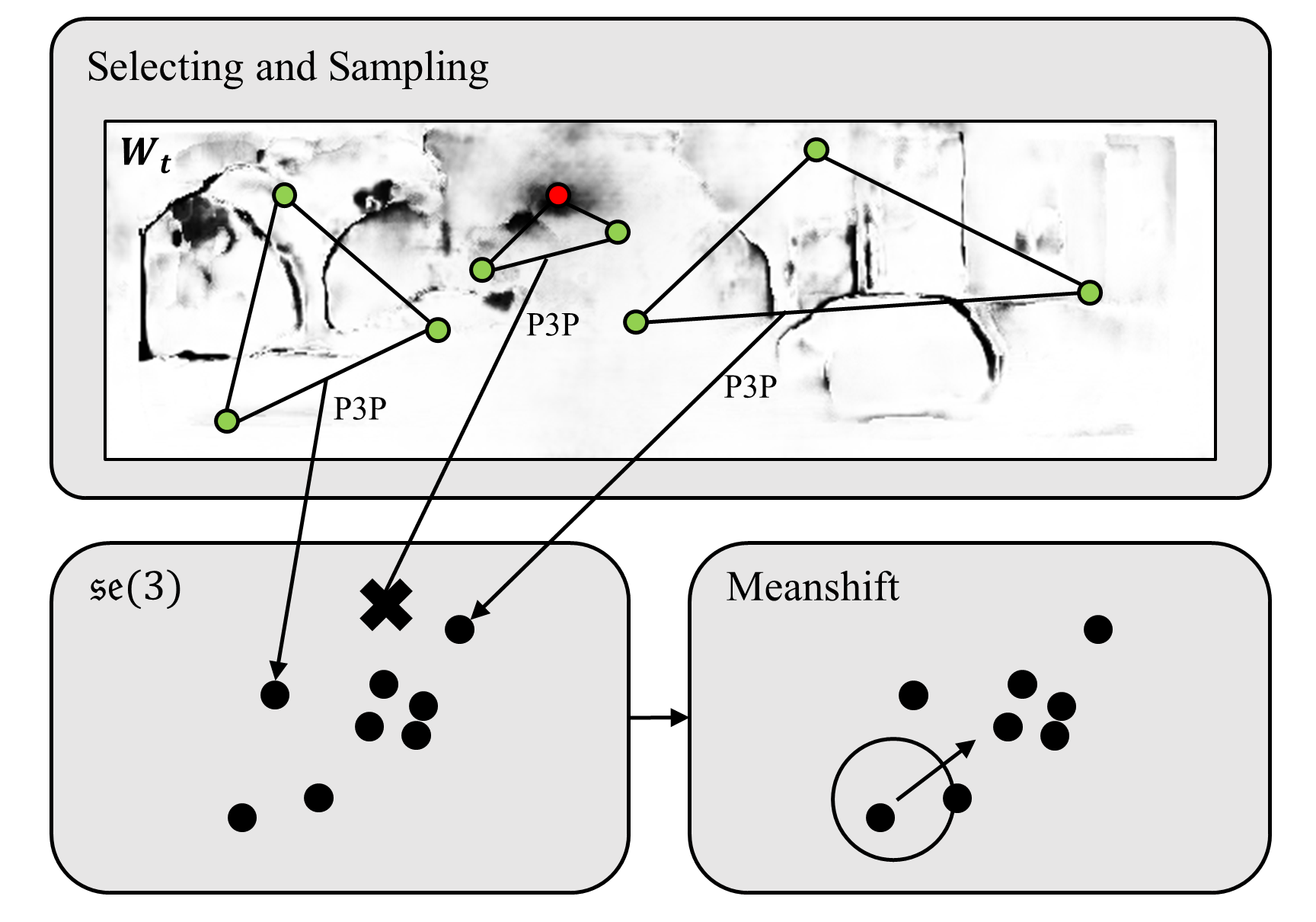}
\caption{\small \textbf{Pose MAP approximation via meanshift-based mode search.}
Each 3D-2D correspondence is part of a unique minimal P3P instance, constituting a pose sample that is weighted by the rigidness map. We map samples to $\mathfrak{se}(3)$ and run meanshift to find the mode.
}
\label{fig:workflow_infer_pose}
\end{figure}

We have approximated the posterior $P(\mathbb{T} \mid \mathbb{X} \,;\, \boldsymbol{\theta}, \mathbb{W})$ with a weighted combination of $q(\boldsymbol{T_t^g})$. Solving for the optimum $\boldsymbol{T_t^*}$ on the posterior equates to finding the mode of the posterior. Since we assume all $q(\boldsymbol{T_t^g})$ to share the same covariance structure, mode finding on this distribution equates to applying meanshift \cite{cheng1995mean} with a Gaussian kernel of covariance $\boldsymbol{\Sigma}$. Note that since $\boldsymbol{\hat{T}_t^g}$ lies in $SE(3)$ \cite{blanco2010tutorial}, while meanshift is applied to vector space, an obtained a mode can not be guaranteed to lie in $SE(3)$. Thus, poses $\boldsymbol{\hat{T}_t^g}$ are first converted to a 6-vector $\boldsymbol{p}=log(\boldsymbol{\hat{T}_t^g})\in \mathfrak{se}(3)$ in Lie algebra through logarithm mapping, and meanshift is applied to the 6-vector space.

\subsection{Algorithm Integration} \label{sec:algorithm_integration}

We now describe the integrated workflow of our visual odometry algorithm, which we denote \name{}. Per Table \ref{table:algorithm}, our input is a sequence of dense optical flows $\mathbb{X}=\{ \boldsymbol{X_t} \mid t=1 \cdots t_N \}$, and our output will be the camera poses of each frame $\mathbb{T}=\{ \boldsymbol{T_t} \mid t=1 \cdots t_N \}$ as well as the depth map $\boldsymbol{\theta}$ of the first frame. Usually, 4-8 optical flows per batch are used. Firstly, \name{}  initializes all $\mathbb{W}$  to one and $\boldsymbol{T_1}$ is initialized from epipolar geometry estimated from $\boldsymbol{X_1}$ using a least median of squares estimator \cite{rousseeuw1984least} 
or, alternatively, from previous estimates if available (i.e. overlapping  consecutive frame batches). 
Then, $\boldsymbol{\theta}$ is obtained from two-view triangulation using $\boldsymbol{T_1}$ and $\boldsymbol{X_1}$. Next, the optimization loop between camera pose, depth map and rigidness map runs until convergence, usually  within 3 to 5 iterations. 
Note we did not smooth the rigidness map before updating camera poses to prevent loss of fine details indicating the potential high-frequency noise in the observations.

\begin{table}[]
\centering
\begin{tabular}{l}
\hline
\textbf{Input}: Optical flow sequence $\mathbb{X}=\{ \boldsymbol{X_t} \mid t=1 \cdots t_N \}$ \\ \hline
\textbf{Output}: Camera poses $\mathbb{T}=\{ \boldsymbol{T_t} \mid t=1 \cdots t_N \}$\\
\qquad \quad \;\,  Depth map $\boldsymbol{\theta}$ of the first frame \\ \hline
Initialize $\mathbb{W}=\{\boldsymbol{W_t} \mid t=1 \cdots t_N \}$ all to one\\
Initialize $\boldsymbol{T_1}$ using epipolar geometry from $\boldsymbol{X_1}$\\
Triangulate $\boldsymbol{\theta}$ from $\boldsymbol{T_1}$ and $\boldsymbol{X_1}$\\
Repeat until $\mathbb{T}$ converges \\
\qquad For $i=1 \cdots t$ \\
\qquad \qquad Update $\boldsymbol{T_i}$ according to Eq. \eqref{eq:pose_criteria}\\
\qquad Update and smooth $\mathbb{W}$ according to Eq. \eqref{eq:rigidness_smooth}\\
\qquad Update $\boldsymbol{\theta}$ according to Eq. \eqref{eq:depth_criteria}\\
\qquad Update $\mathbb{W}$ according to Eq. \eqref{eq:rigidness_smooth} w/o smoothing \\ \hline
\end{tabular}
\caption{\small  \textbf{The \name{} algorithm.}}
\label{table:algorithm}
\end{table}

\section{Experiments}

\begin{figure*}[]
\centering
\includegraphics[width=0.49\columnwidth]{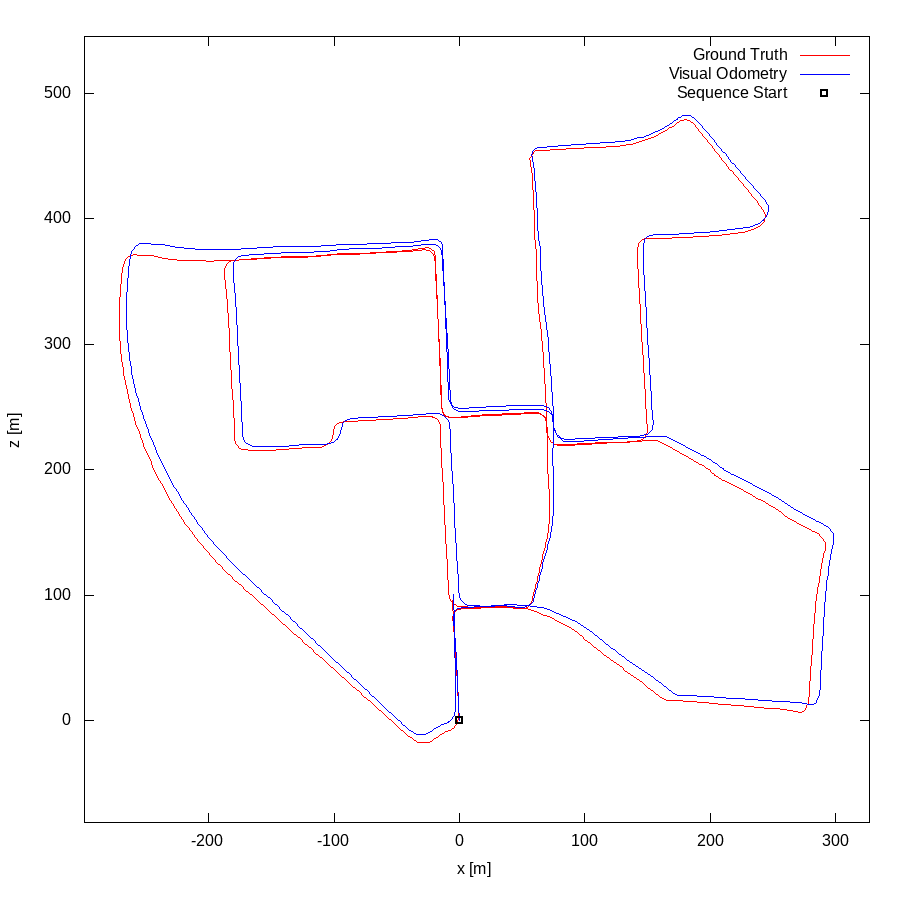}
\includegraphics[width=0.49\columnwidth]{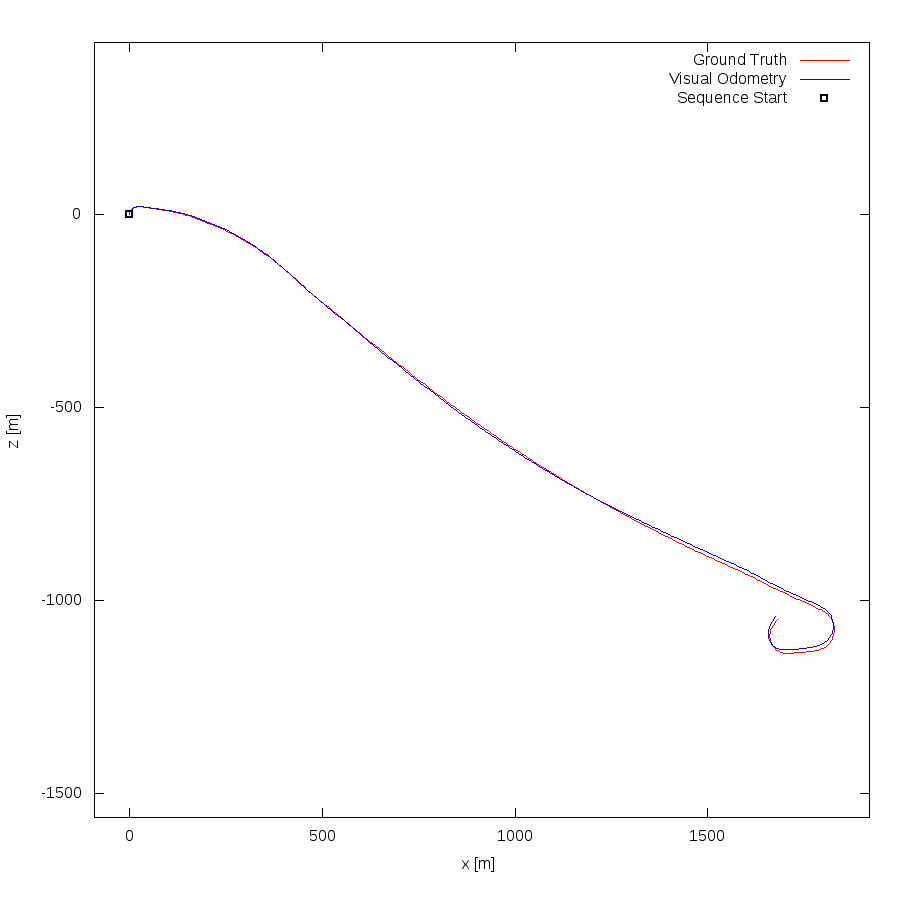}
\includegraphics[width=0.49\columnwidth]{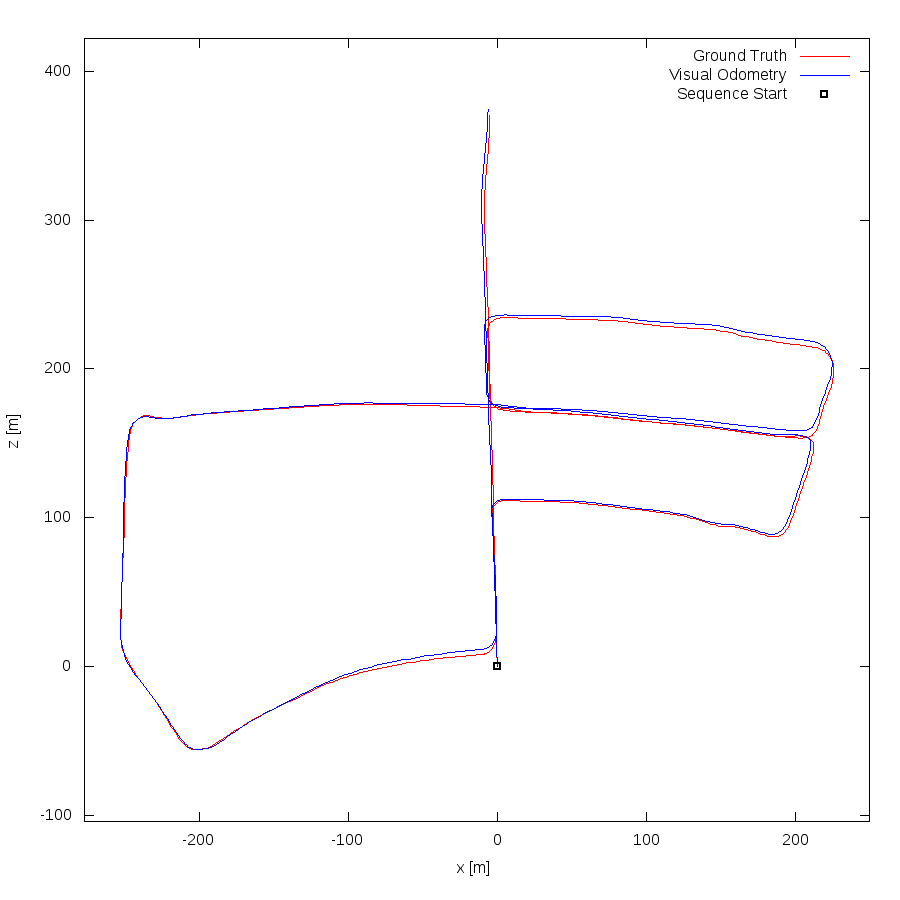}
\includegraphics[width=0.49\columnwidth]{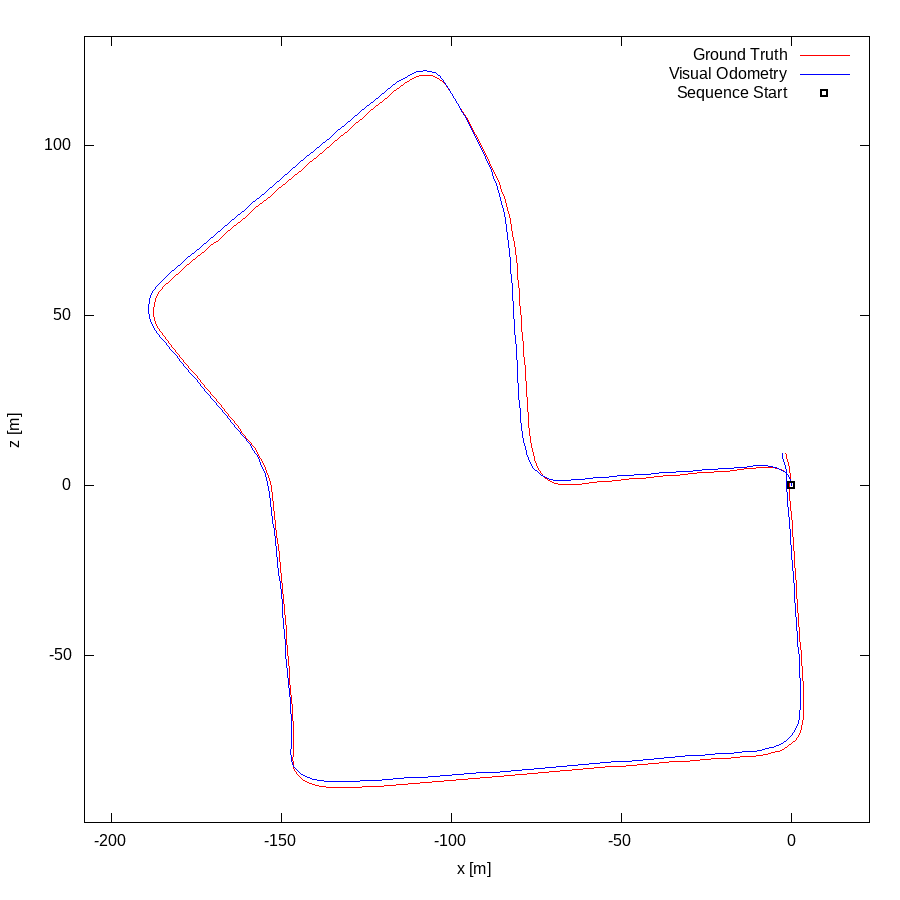}
\caption{\small \textbf{Results on KITTI sequences 00,\,01,\,05,\,07}. Monocular \name{} does not deploy any mapping, bundle adjustment or loop closure. Scale is estimated assuming fixed and known camera height from the ground.}
\label{fig:kitti_train}
\end{figure*}

\subsection{KITTI Benchmark} \label{sec:kitti_evaluation}

We tested on the KITTI odometry benchmark \cite{geiger2012we} 
of a car driving at urban and highway environments. We use PWC-Net \cite{sun2018pwc} as our external dense optical flow input. The sliding window size is set to 6 frames. We set $\lambda$ in Eq. \eqref{eq:adaptive_uniform} to 0.15, $\gamma$ in Eq. \eqref{eq:smoothness} to 0.9. The Gaussian kernel covariance matrix $\boldsymbol{\Sigma}$ in Eq. \eqref{eq:pose_variational_distribution} is set to diagonal, scaled to 0.1 and 0.004 at the dimensions of translation and rotation respectively.
The hyper-parameters for the Fisk residual model are $a_1=0.01,\,a_2=0.09,\,b_1=-0.0022,\,b_2=1.0$, obtained from Figure \ref{fig:fisk_1}. Finally, absolute scale is estimated from ground plane by taking the mode of pixels with surface normal vector near perpendicular. More details of ground plane estimation are provided in the appendix.

\begin{table}[]
\centering
\resizebox{0.9\columnwidth}{!}{%
\begin{tabular}{|c|cc|cc|cc|}
\hline
& \multicolumn{2}{c|}{VISO2-M} & \multicolumn{2}{c|}{MLM-SFM} & \multicolumn{2}{c|}{\name{}} \\ \hline
Sequence & \begin{tabular}[c]{@{}c@{}}Trans.\\ (\%)\end{tabular} & \begin{tabular}[c]{@{}c@{}}Rot.\\ (deg/m)\end{tabular} & \begin{tabular}[c]{@{}c@{}}Trans.\\ (\%)\end{tabular} & \begin{tabular}[c]{@{}c@{}}Rot.\\ (deg/m)\end{tabular} & \begin{tabular}[c]{@{}c@{}}Trans.\\ (\%)\end{tabular} & \begin{tabular}[c]{@{}c@{}}Rot.\\ (deg/m)\end{tabular} \\ \hline
00 & 12.53 & 0.0260 & 2.04 & 0.0048 & \textbf{1.09} & \textbf{0.0039} \\
01 & 28.09 & 0.0641 & - & - & \textbf{2.31} & \textbf{0.0037} \\
02 & 3.98 & 0.0123 & 1.50 & \textbf{0.0035} & \textbf{1.19} & 0.0042 \\
03 & 4.09 & 0.0206 & 3.37 & \textbf{0.0021} & \textbf{1.46} & 0.0034 \\
04 & 2.58 & 0.0162 & 1.43 & \textbf{0.0023} & \textbf{1.13} & 0.0049 \\
05 & 14.68 & 0.0379 & 2.19 & \textbf{0.0038} & \textbf{1.15} & 0.0041 \\
06 & 6.73 & 0.0195 & 2.09 & 0.0081 & \textbf{1.13} & \textbf{0.0045} \\
07 & 14.95 & 0.0558 & - & - & \textbf{1.63} & \textbf{0.0054} \\
08 & 11.63 & 0.0215 & 2.37 & \textbf{0.0044} & \textbf{1.50} & \textbf{0.0044} \\
09 & 4.94 & 0.0140 & 1.76 & 0.0047 & \textbf{1.61} & \textbf{0.0039} \\
10 & 23.36 & 0.0348 & 2.12 & 0.0085 & \textbf{1.44} & \textbf{0.0043} \\ \hline
Avg. & 10.85 & 0.0249 & 2.03 & 0.0045 & \textbf{1.32} & \textbf{0.0042} \\ \hline
\end{tabular}
}
\caption{\small \textbf{Results on KITTI training sequences 0-10}. The translation and rotation errors are averaged over all sub-sequences of length from 100 meters to 800 meters with 100 meter steps.}
\label{table:kitti_train}
\end{table}

\begin{table}[]
\centering
\resizebox{0.68\columnwidth}{!}{%
\begin{tabular}{|c|cc|}
\hline
Method & Trans. (\%) & Rot. (deg/m) \\ \hline
VISO2-M \cite{geiger2011stereoscan} & 11.94 & 0.0234 \\
MLM-SFM \cite{song2014robust} \cite{song2013parallel} & 2.54 & 0.0057 \\
PbT-M2 \cite{fanani2017predictive} \cite{fanani2016keypoint} \cite{fanani2017multimodal} & 2.05 & 0.0051 \\
BVO \cite{pereira2017backward}  & 1.76 & \textbf{0.0036} \\
\textbf{VOLDOR (Ours)} & \textbf{1.65} & 0.0050 \\ \hline
VO3pt* \cite{alcantarilla2012combining} \cite{alcantarilla2011vision} & 2.69 & 0.0068 \\
VISO2-S* \cite{geiger2011stereoscan} & 2.44 & 0.0114 \\
eVO* \cite{sanfourche2013evo} & 1.76 & 0.0036 \\
S-PTAM* \cite{pire2017s} \cite{pire2015stereo} & 1.19 & 0.0025 \\
ORB-SLAM2* \cite{mur2017orb} & 1.15 & 0.0027 \\ \hline
\end{tabular}
}
\caption{\small \textbf{Results on KITTI odometry testing sequences 11-21.} \; * indicates the method is based on stereo input.}
\label{table:kitti_test}
\end{table}

Table \ref{table:kitti_train} and Figure \ref{fig:kitti_train} are our results on the KITTI odometry training set sequences 0-10. We picked VISO2 \cite{geiger2011stereoscan} and MLM-SFM \cite{song2014robust} \cite{song2013parallel} as our baselines, whose scales are also estimated from ground height. Table \ref{table:kitti_test} compares our result with recent popular methods on KITTI test set sequences 11-21, where we download from KITTI odometry official ranking board. As the results shows, \name{} has achieved top-ranking accuracy under KITTI dataset among monocular methods.

Table \ref{table:kitti_depth} shows our depth map quality on the KITTI stereo benchmark. We masked out foreground moving object as well as aligned our depth with groundtruth to solve scale ambiguity. We separately evaluated the depth quality for different rigidness probabilities, where $W^j=\sum_tW_t^j$. The EPE of PSMNet \cite{chang2018pyramid}, GC-Net \cite{kendall2017end} and GA-Net \cite{Zhang_2019_CVPR} are measured on stereo 2012 test set and background outlier percentage is measured on stereo 2015 test set while our method is measured on training set on stereo 2015.

\begin{table}[]
\centering
\resizebox{0.75\columnwidth}{!}{%
\begin{tabular}{|c|c|c|c|}
\hline
Methods & Density & EPE / px & bg-outlier \\ \hline
GC-Net \cite{kendall2017end} & \textbf{100.00\%} & 0.7 & 2.21\% \\
PSMNet \cite{chang2018pyramid} & \textbf{100.00\%} & 0.6 & 1.86\% \\
GA-Net-15 \cite{Zhang_2019_CVPR} & \textbf{100.00\%} & \textbf{0.5} & \textbf{1.55}\% \\ \hline
Ours $(W^j>5)$ & 27.07\% & \textbf{0.5616} & \textbf{1.47\%} \\
Ours $(W^j>4)$ & 37.87\% & 0.6711 & 2.05\% \\
Ours $(W^j>3)$ & 49.55\% & 0.7342 & 2.56\% \\
Ours $(W^j>2)$ & 62.50\% & 0.8135 & 3.17\% \\
Ours $(W^j>1)$ & 78.18\% & 0.9274 & 4.17\% \\
Ours $(W^j>0)$ & \textbf{100.00\%} & 1.2304 & 5.82\% \\ \hline
\end{tabular}
}
\caption{\small \textbf{Results on KITTI stereo benchmark.} A pixel is considered as outlier if disparity EPE is $>3px$ and $>5\%$. $W^j$ denotes the sum of pixel rigidness $W^j=\sum_tW_t^j$.}
\label{table:kitti_depth}
\end{table}

\begin{table}[]
\centering
\resizebox{0.88\columnwidth}{!}{%
\begin{tabular}{|c|c|c|c|c|}
\hline
Sequence & \begin{tabular}[c]{@{}c@{}}ORB-SLAM2\\ (RGB-D)\end{tabular} & \begin{tabular}[c]{@{}c@{}}DVO-SLAM\\ (RGB-D)\end{tabular} & \begin{tabular}[c]{@{}c@{}}DSO\\ (Mono)\end{tabular} & \begin{tabular}[c]{@{}c@{}}Ours\\ (Mono)\end{tabular} \\ \hline
fr1/desk & 0.0163 & 0.0185 & 0.0168 & \textbf{0.0133} \\
fr1/desk2 & 0.0162 & 0.0238 & 0.0188 & \textbf{0.0150} \\
fr1/room & 0.0102 & 0.0117 & 0.0108 & \textbf{0.0090} \\
fr2/desk & \textbf{0.0045} & 0.0068 & 0.0048 & 0.0053 \\
fr2/xyz & 0.0034 & 0.0055 & \textbf{0.0025} & 0.0034 \\
fr3/office & 0.0046 & 0.0102 & 0.0050 & \textbf{0.0045} \\
fr3/nst & 0.0079 & 0.0073 & 0.0087 & \textbf{0.0071} \\ \hline
\end{tabular}
}
\caption{\small \textbf{Results on TUM RGB-D dataset.} The values are translation RMSE in meters.}
\label{table:tum_data}
\end{table}
\begin{figure*}[]
\centering
\begin{tabular}{|cc|cc|}
\hline
\subfloat[\scriptsize{Depth likelihood with Gaussian-(MLE/MIE) Residual Model}]{\includegraphics[width=0.46\columnwidth]{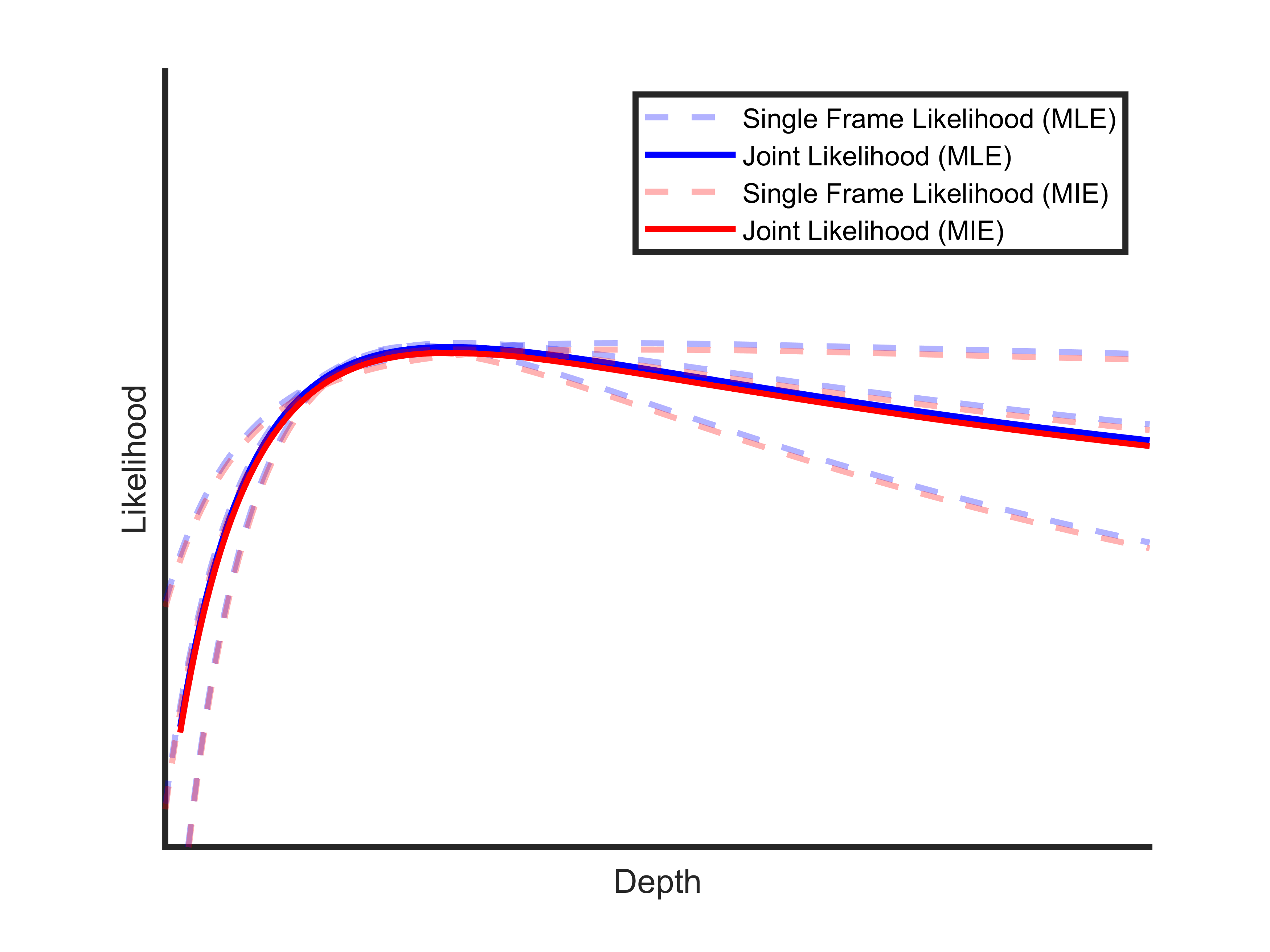}} & 
\subfloat[Epipole distribution with Gaussian Residual Model]{\includegraphics[width=0.5\columnwidth]{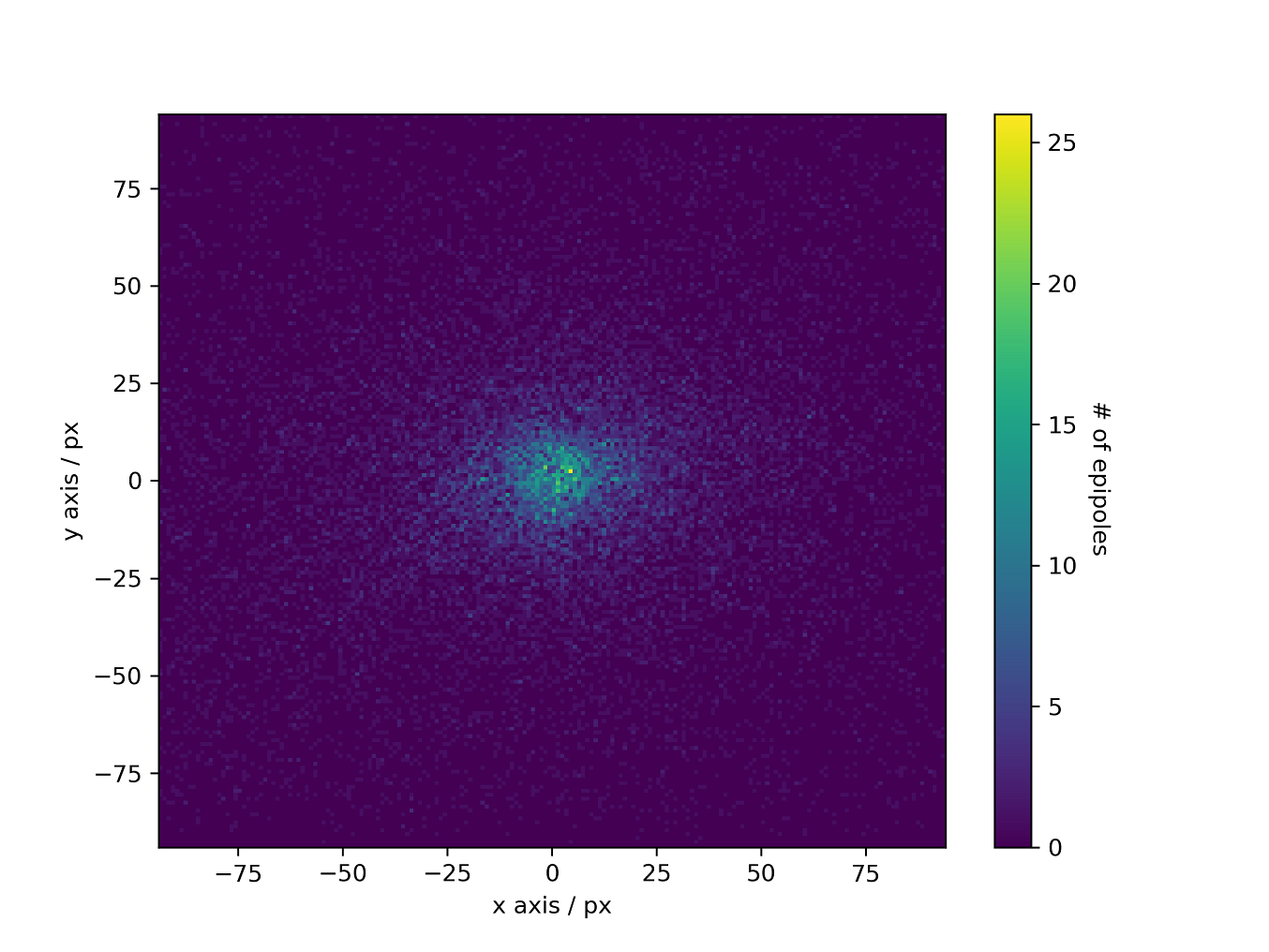}} &
\subfloat[\scriptsize{Depth likelihood with Fisk-(MLE/MIE) Residual Model}]{\includegraphics[width=0.46\columnwidth]{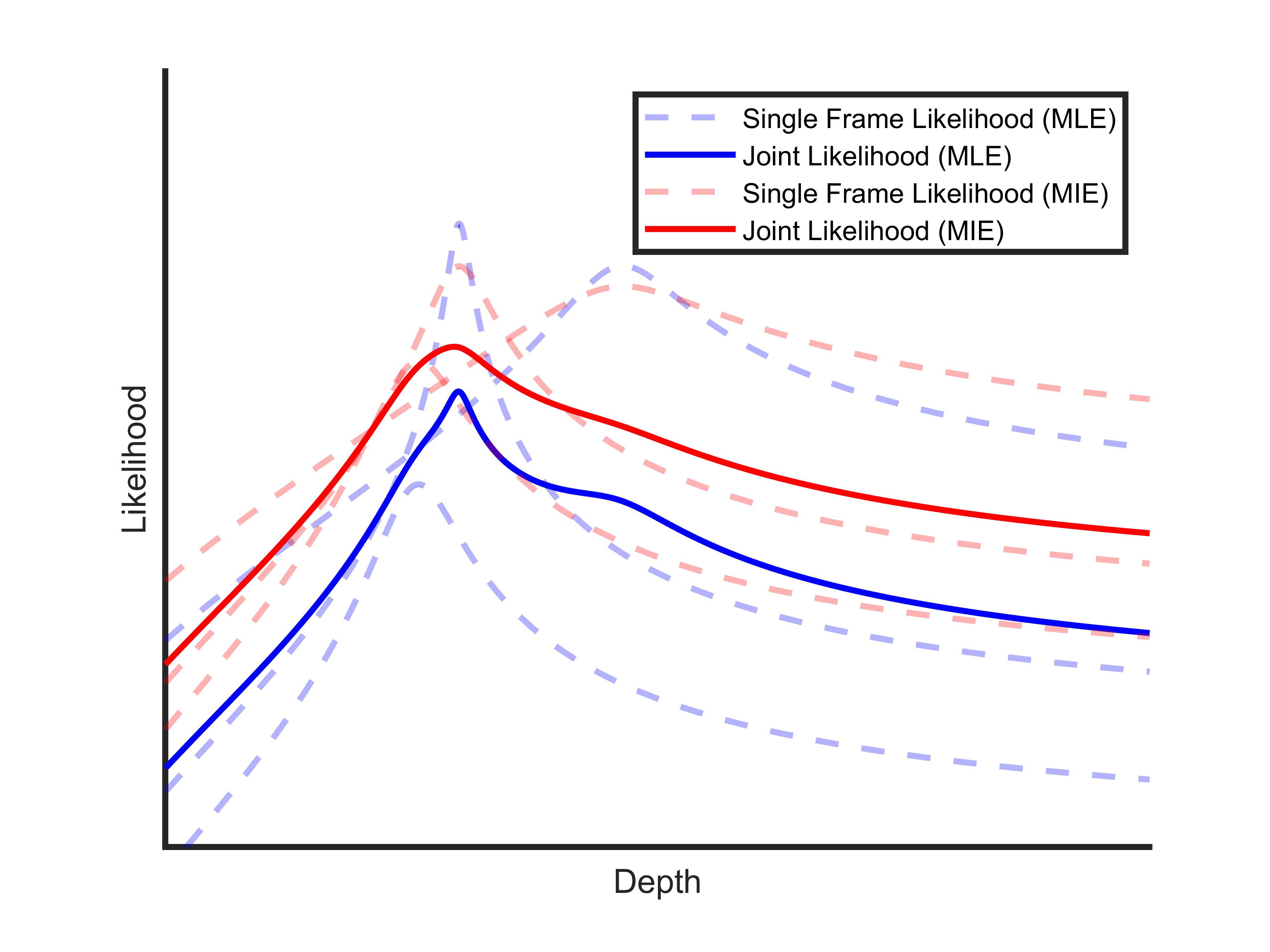}} & 
\subfloat[Epipole distribution with Fisk Residual Model]{\includegraphics[width=0.5\columnwidth]{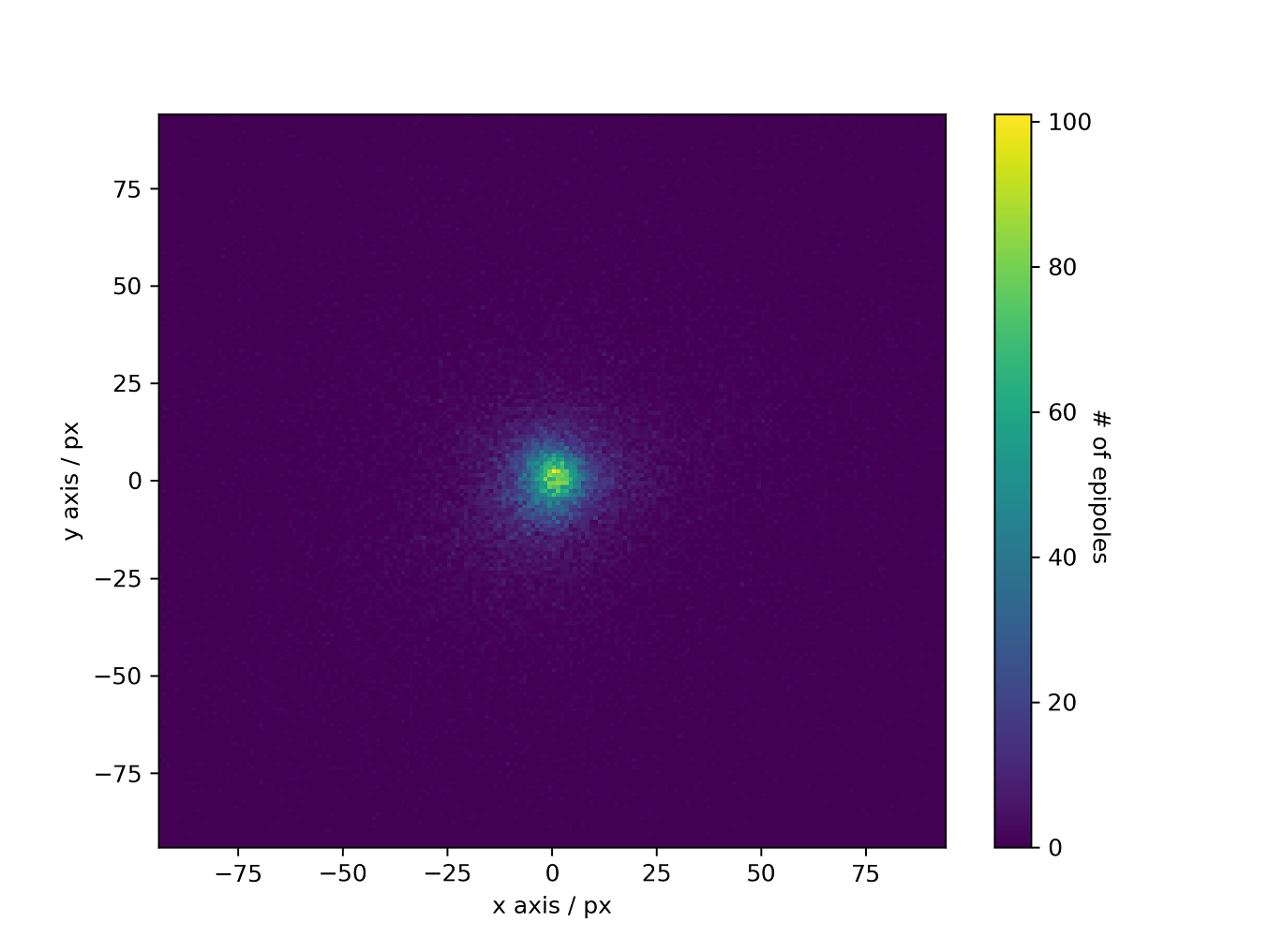}} \\ \hline
\end{tabular}
\caption{\small \textbf{Fisk model qualitative study.} (a) and (c) visualize the depth likelihood function under Gaussian and Fisk residual models with MLE and MIE criteria. Dashed lines indicate the likelihood given by a single optical flow. Solid lines are joint likelihood obtained by fusing all dashed lines. MLE and MIE are shown in different colors. (c) and (d) visualize the epipole distribution for 40K camera pose samples. For better visualization, the density color bars of (b) (d) are scaled  differently.}
\label{fig:fisk_goodness}
\begin{tabular}{|cc|cc|}
\hline
\subfloat[Ablation study over camera pose]{\includegraphics[width=0.48\columnwidth]{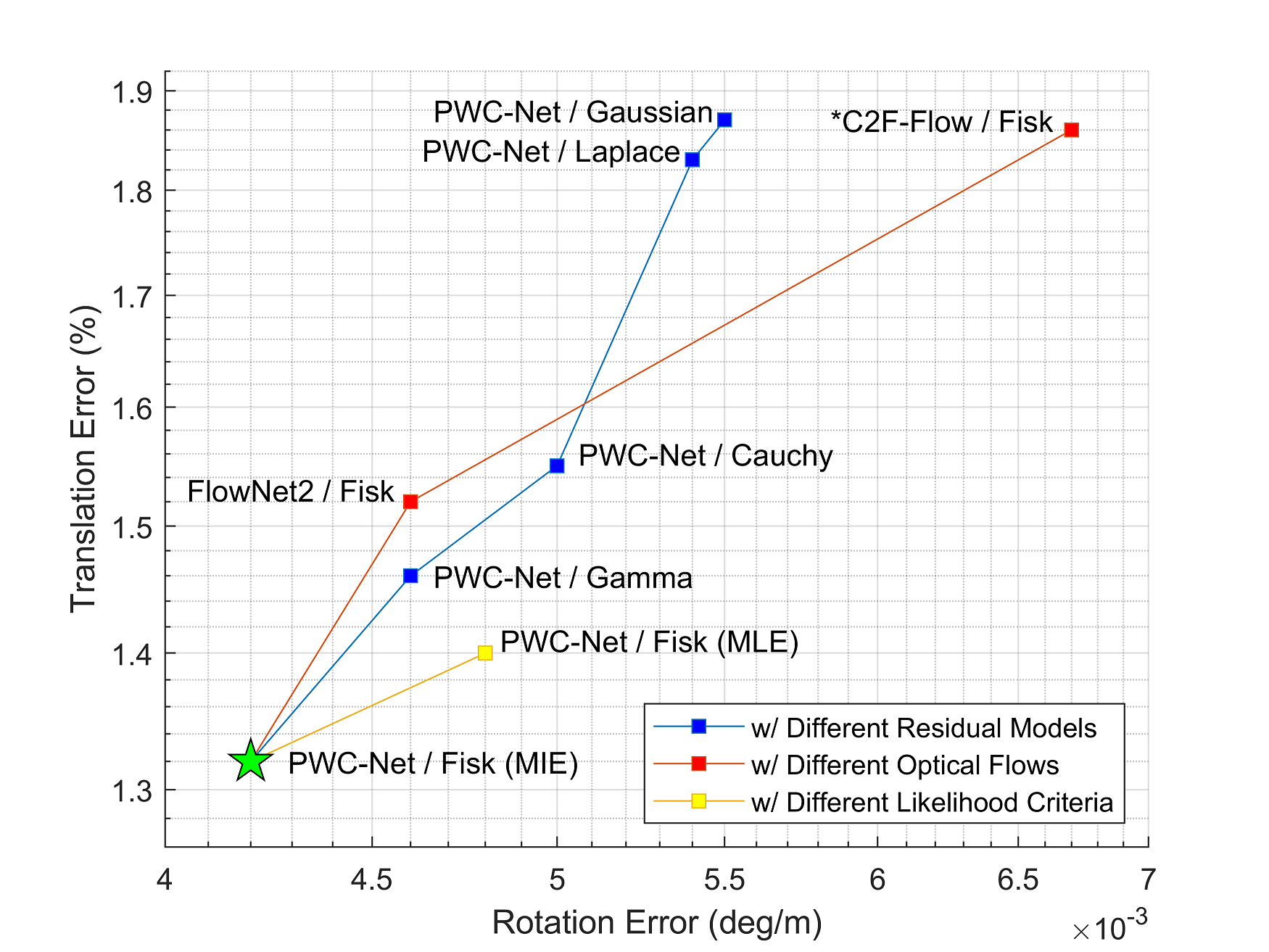}} & 
\subfloat[Ablation study over depth]{\includegraphics[width=0.48\columnwidth]{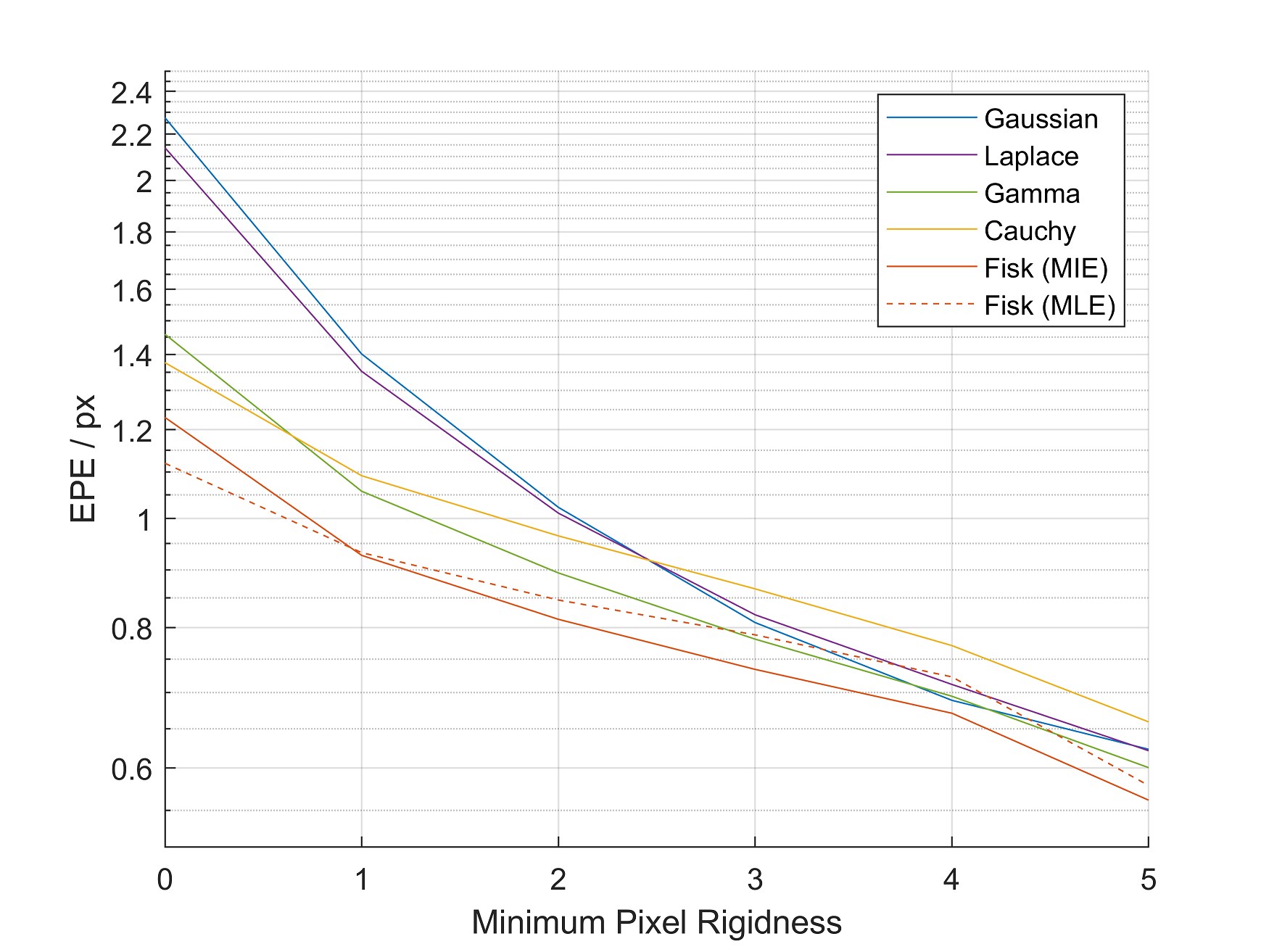}} &
 \subfloat[Timing over frame numbers]{\includegraphics[width=0.48\columnwidth]{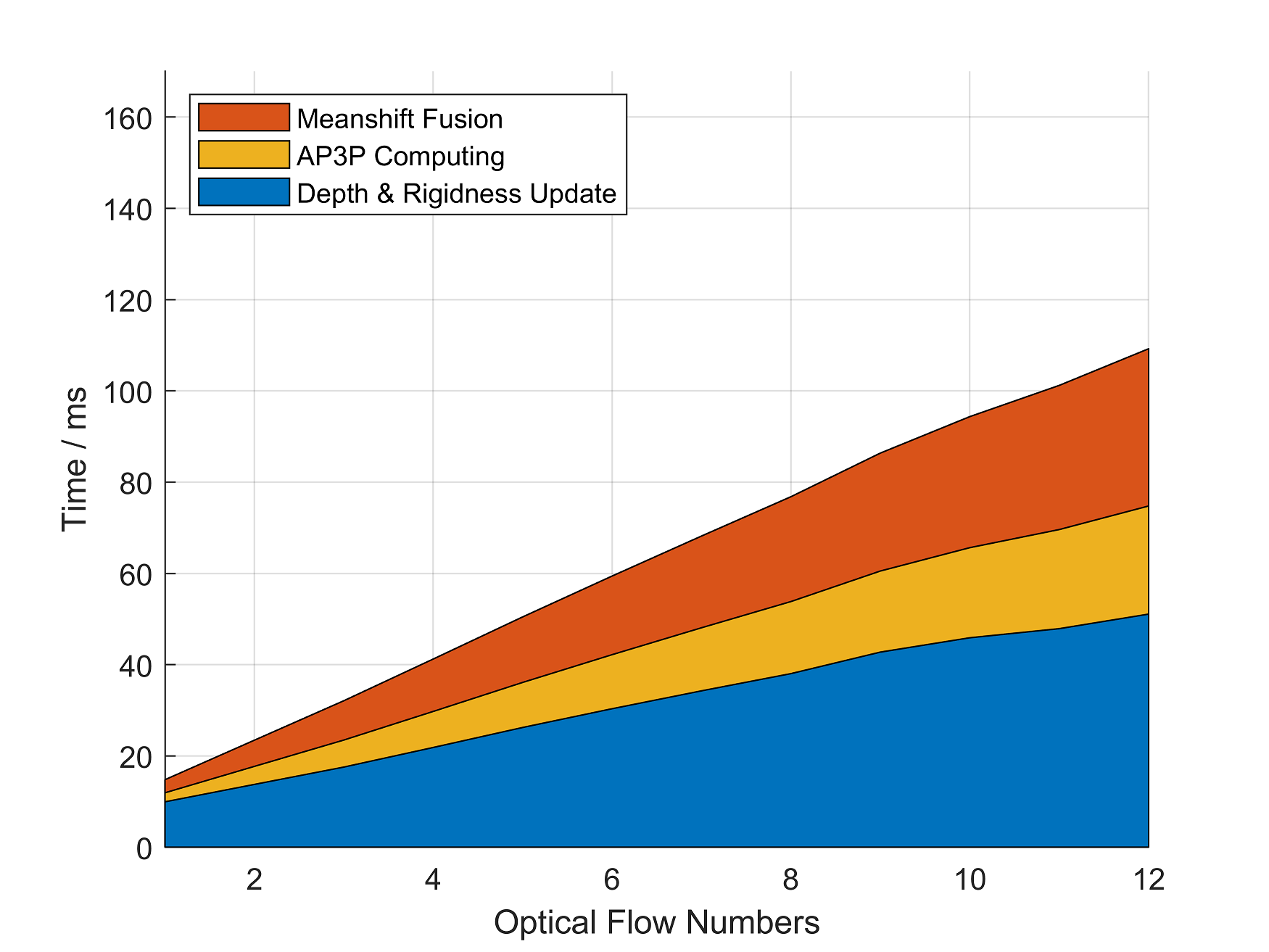}} & 
\subfloat[Timing over pose samples]{\includegraphics[width=0.48\columnwidth]{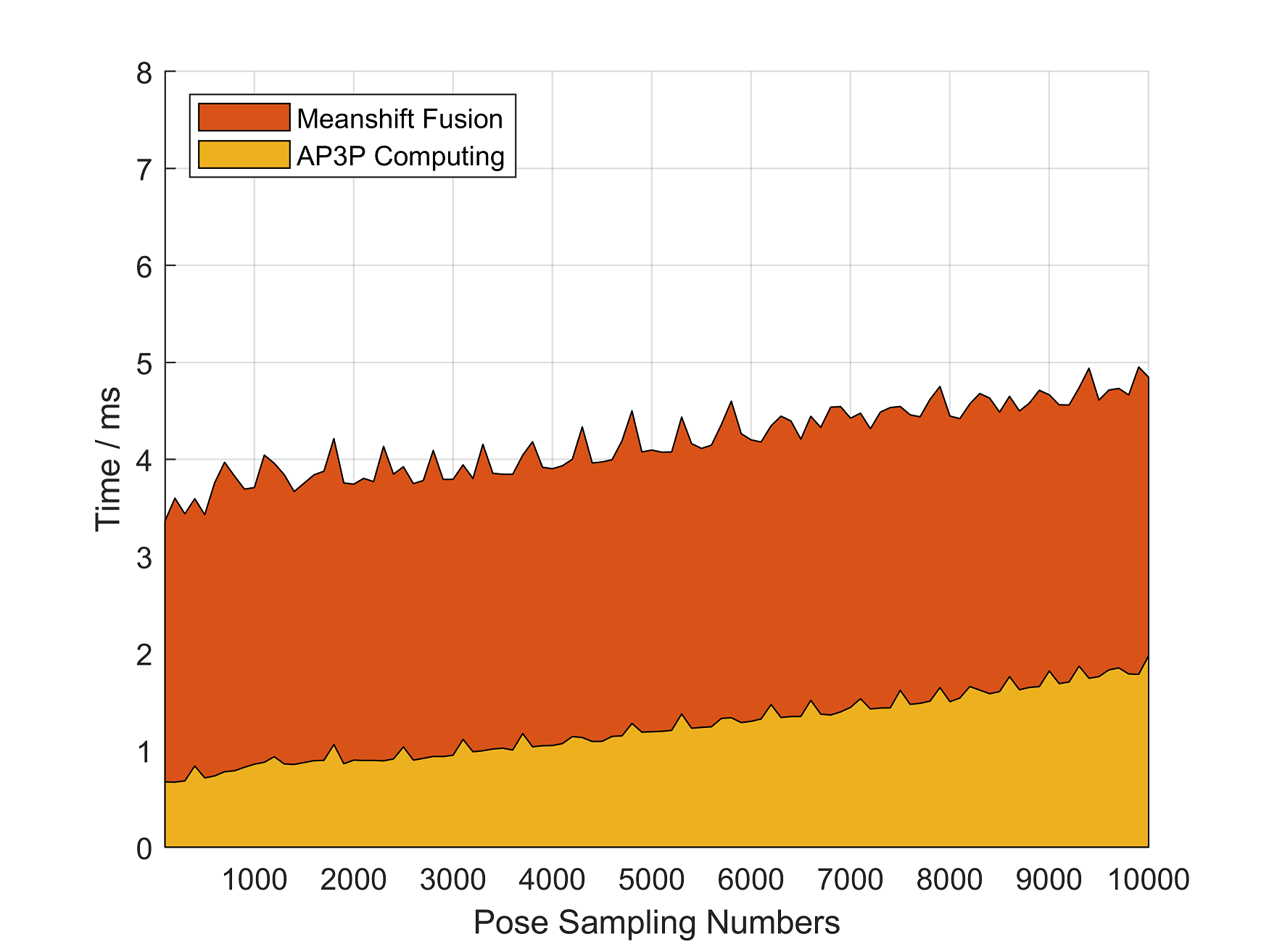}}
\\ \hline
\end{tabular}
\caption{\small \textbf{Ablation study and runtime.} (a) shows the camera pose error of \name{} under different residual models and dense optical flow input (*due to noisy ground estimations given by C2F-Flow, its scale is corrected using groundtruth). (b) shows our depth map accuracy under different residual models. (c) and (d) show the runtime of our method tested on a GTX 1080Ti GPU.}
\label{fig:ablation_and_timing}
\end{figure*}


\subsection{TUM RGB-D Benchmark} \label{sec:tum_evaluation}
Accuracy experiments on TUM RGB-D \cite{sturm2012benchmark} compared \name{} vs. full SLAM systems. 
In all instances, we rigidly align trajectories to groundtruth for segments with 6 frames and estimate mean translation RMSE of all segments. Parameters remains the same to KITTI experiments. Our comparative baselines are an indirect sparse method ORB-SLAM2 \cite{mur2017orb}, a direct sparse method DSO \cite{engel2017direct} and a dense direct method DVO-SLAM \cite{kerl2013dense}. Per Table \ref{table:tum_data}, \name{} performs well under indoor capture exhibiting smaller camera motions and diverse motion patterns.

\subsection{Ablation and Performance Study} \label{sec:ablation_exp}

Figure \ref{fig:fisk_goodness}  visualizes the depth likelihood function and camera pose sampling distribution. 
With our Fisk residual model, depth likelihood from each single frame has a well localized extremum (Fig.\ref{fig:fisk_goodness}-c), compared to a Gaussian residual model (Fig.\ref{fig:fisk_goodness}-a). This leads to a joint likelihood having a more distinguishable optimum, and results in more concentrated camera pose samplings (Fig.\ref{fig:fisk_goodness}-b,d). Also, with the MIE criteria, depth likelihood of the Fisk residual model is relaxed to a smoother shape whose effectiveness is further analyzed in Fig. \ref{fig:ablation_and_timing}, while a Gaussian residual model is agnostic to the choice between MIE and MLE (Fig.\ref{fig:fisk_goodness}-a).
Per the quantitative study in Fig. \ref{fig:ablation_and_timing} (b), compared to other analytic distributions, the Fisk residual model gives significantly better depth estimation when only a small number of reliable observations (low $W^j$) are used. 
Performance across different residual models tends to converge as the number of reliable samples increases (high $W^j$), while the Fisk residual model still provides the lowest EPE.
Figure \ref{fig:ablation_and_timing} (a) shows the ablation study on camera pose estimation for three optical flow methods, four residual models and our proposed MIE criteria. The accuracy of combining PWC-Net optical flow, a Fisk residual model, and our MIE criteria, strictly dominates (in the Pareto sense) all other combinations.
Figure \ref{fig:ablation_and_timing} (b) shows the MIE criteria yields depth estimates that are more consistent across the entire image sequence, leading to improved overall accuracy. However, at the extreme case of predominantly unreliable observations (very low $W^j$) MLE provides the most accurate depth.
Figure \ref{fig:ablation_and_timing} (c) shows the overall runtime evaluation for each component under different frame numbers. Figure \ref{fig:ablation_and_timing} (d) shows runtime for pose update under different sample rates.

\vspace{-0.05cm}
\section{Conclusion}
\vspace{-0.05cm}
Conceptually, we pose the VO problem as an instance of geometric parameter inference under the supervision of an adaptive model of the empirical distribution of dense optical flow residuals. Pragmatically, we develop a monocular VO pipeline which obviates the need for a)  feature extraction, b) ransac-based estimation, and c) local bundle adjustment, yet still achieves top-ranked performance in the KITTI and TUM RGB-D benchmarks. We posit the use of dense-indirect representations and adaptive data-driven supervision as a general and extensible framework for multi-view geometric analysis tasks.


{\small
\bibliographystyle{ieee_fullname}
\bibliography{egbib}
}

\end{document}